\documentclass[11pt]{article}

\usepackage[preprint]{acl}

\usepackage{times}
\usepackage{latexsym}

\usepackage[T1]{fontenc}

\usepackage[utf8]{inputenc}

\usepackage{microtype}

\usepackage{inconsolata}

\usepackage{graphicx}

\usepackage{booktabs}
\usepackage{adjustbox}
\usepackage{colortbl}
\definecolor{band}{gray}{0.92}
\usepackage{amsmath}
\usepackage{amssymb}

\title{Where Do Distilled Students Store Thinking?}

\author{
  \textbf{Taeyeong Kim\textsuperscript{1}},
  \textbf{Jaemin Jung\textsuperscript{1}},
  \textbf{Ahhyun Kim\textsuperscript{1}},
  \textbf{Unggi Lee\textsuperscript{2*}}
\\
\\
  \textsuperscript{1}Chosun University,
  \textsuperscript{2}Korea University Sejong Campus
\\
  \small{
    \textsuperscript{*}Corresponding author: \href{mailto:codingchild@korea.ac.kr}{codingchild@korea.ac.kr}
  }
}

\begin{document}
\maketitle
\begin{abstract}
Small students degrade on large reasoning models' (LRMs') long solutions, so recent work shortens them by interpolating an LRM with a short-answer model at ratio $\lambda$. We ask what student weights retain when this teacher stops thinking, comparing LoRA adapters of three students from three teacher pairs at each $\lambda$. In each pair, the adapters' update subspace shifts most where the teacher stops thinking and solutions shorten most, locating the change without attributing it. We identify a shared adapter subspace tied to thinking-like behavior (reflective markers, monologue format and long solutions). Shared by adapters before that point, it keeps about half (post hoc) while one pair's teacher halves its solutions but keeps its think tag, and largely empties after thinking stops. Removing it largely turns this behavior off for every pair, student and benchmark. Only in that pair and one other student does it leave clearly fewer markers than removing the update's top components; elsewhere, same-energy random directions often nearly match it. Grafted into a short-answer student at twice its weight, it turns early markers (a post hoc count) on, like the whole update, but the student rarely finishes. Our work identifies a weight subspace tied to distilled thinking-like behavior.
\end{abstract}

\section{Introduction}
\label{sec:intro}

\begin{figure*}[t]
\centering
\includegraphics[width=0.82\textwidth]{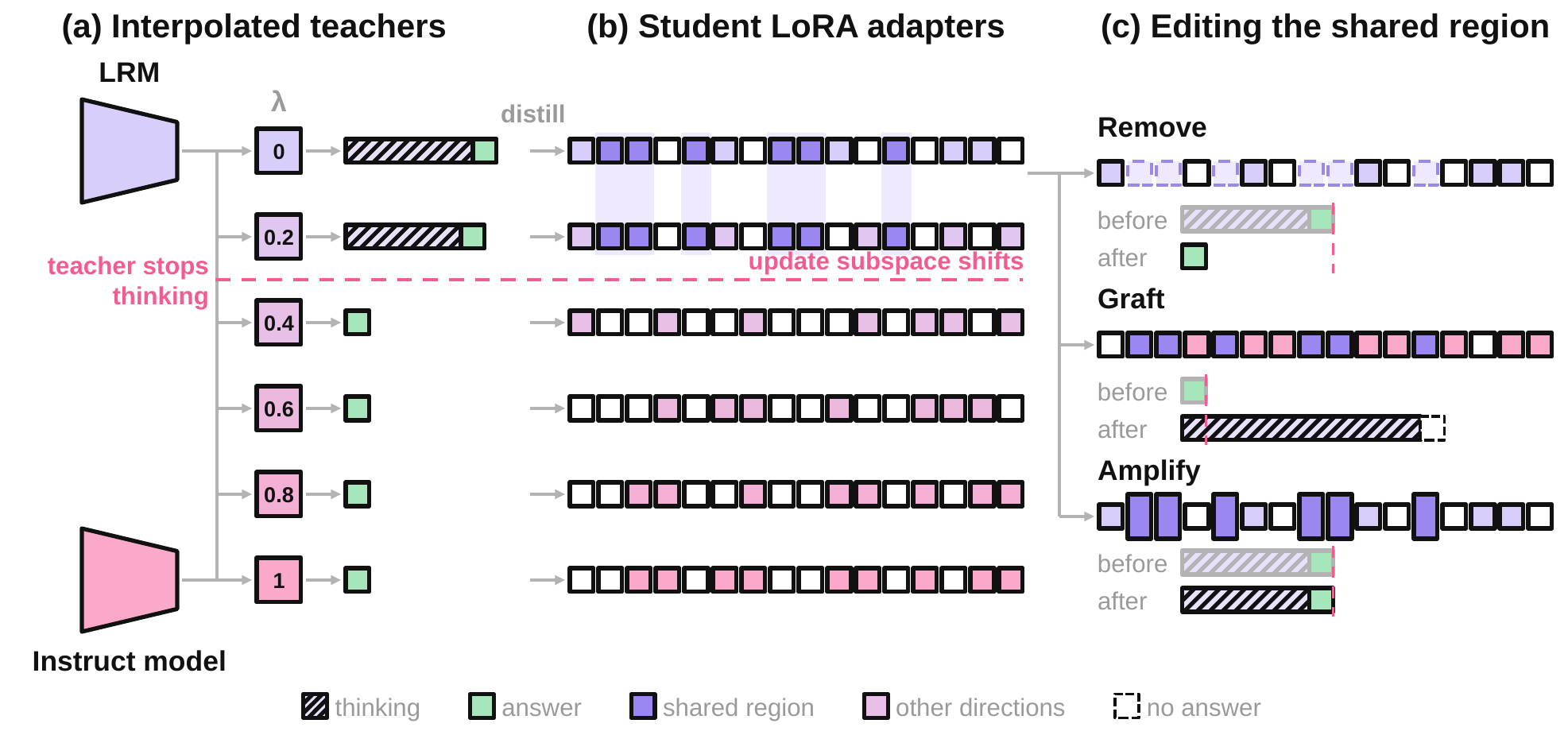}
\caption{(a) Teachers interpolate an LRM ($\lambda=0$) and an instruct model ($\lambda=1$) and stop thinking past some $\lambda$ (dashed for pair A, thinking hatched). (b) Pre-switch adapters share 5--8 (weighted) of 16 directions (purple), which later adapters largely leave. (c) Removing the region from the $\lambda=0$ adapter leaves mostly the answer, grafting it into the $\lambda=1$ adapter at twice its weight turns early markers on but rarely finishes, and amplifying it 1.2--1.6$\times$ at a fixed norm moves early markers at most 1.8$\times$. Light bars precede each edit, and lengths are schematic.}
\label{fig:overview}
\end{figure*}

Distilling the long reasoning of large reasoning models (LRMs) into small models has become widely used \citep{guo2025deepseekr1,muennighoff2025s1}. Before answering, an LRM writes a long reflective monologue, which we call thinking. However, small students perform worse when distilled from these long solutions \citep{li2025small,luo2025valley}. Recent work adjusts solution length with an interpolated teacher, which interpolates the weights of an LRM ($\lambda=0$) and a short-answer instruct model ($\lambda=1$) with ratio $\lambda$ \citep{lan2026midistillation}.
As $\lambda$ grows, such a teacher almost entirely stops thinking once $\lambda$ exceeds a pair-dependent threshold \citep{wu2026interpolation}, and accuracy and token use change abruptly within a narrow $\lambda$ range \citep{lan2025spectrum}. One study reports no such switch \citep{yang2026system}, and these studies analyze the teacher, not the student's weights. Studies of the weights or internals of distilled models compare each distilled model with its own original model \citep{ward2025rank1,hu2026transcoder,venhoff2026base}. Thinking and length fall together in the teacher's outputs \citep{wu2026interpolation} and in the student's (\S\ref{sec:outputs}), so outputs alone cannot tell which one the student learned.

We therefore ask what remains in the student's weights when the interpolated teacher stops thinking.
Full fine-tuning spreads its changes over all weights, whereas low-rank adaptation (LoRA) gathers them in a few update directions that can be compared across runs \citep{hu2021lora} and still captures most of a reasoning fine-tune's gain \citep{ward2025rank1}. We use LoRA adapters as a measuring instrument. For three teacher pairs (A, B and C) and three students, nine cells in all, we compare adapters trained while varying only $\lambda$ against adapters that differ only in their data, and test the subspace we find by editing it (Figure~\ref{fig:overview}).
We identify a shared adapter subspace, the region, tied to thinking-like behavior (reflective markers, monologue format and long solutions). At our eigenvalue cut it spans 5--8 of the 16 update directions (norm-weighted), and its unweighted build holds 47--72\% of the update's energy (Table~\ref{tab:anatomy}). In pair A, training solutions halve while the teacher keeps its think tag and format but not its marker rate. The region keeps about half over this interval under post hoc references, and editing the region switches both format and hesitation in most cells. This gives a weight-level account of what distillation from an interpolated teacher transfers, which output-level studies do not give. Our contributions are as follows.

\begin{itemize}
  \item In each pair, the adapters' update subspaces shift most where the teacher stops thinking and solutions shorten most, locating the change without attributing it, and the region shared before it largely empties later (\S\ref{sec:switch}).
  \item Removing the region largely turns thinking off in all nine cells, more than removing other directions and, in pair A and C/Qwen, the update's top components, while grafting it at twice its weight turns early reflective markers on but rarely yields finished solutions (\S\ref{sec:carries}).
  \item Removal also turns thinking off on GSM8K and GPQA-Diamond and with a region found from two $\lambda=0$ seeds alone, yet no edit beats the untrained student's mean MATH-500 accuracy (\S\ref{sec:general}).
\end{itemize}

\section{Related Work}\label{sec:related}

\subsection{Chain-of-Thought Distillation to Small Students}

Knowledge distillation trains a small student on a strong teacher's outputs \citep{hinton2015distilling}, and language-model students are often trained on sequences the teacher generates \citep{kim2016sequence}. Chain-of-thought distillation transfers a large model's reasoning steps in this way \citep{ho2023reasoning,magister2023teaching}, and distilling the long reasoning trajectories of strong LRMs such as R1 into small models has become the main approach \citep{guo2025deepseekr1}.
LRMs overthink easy problems \citep{chen2025overthinking}. Small students learn better from short solutions \citep{li2025small}, degrade sharply on even a little long data \citep{luo2025valley}, and do not always gain from the strongest teacher \citep{hu2026strongest}.
Past work has combined long and short solutions with length and difficulty curricula \citep{jiang2025teach}, with long2short training \citep{kimi2025k15} and with one teacher's reasoning mode turned on and off \citep{boizard2026scale}. Interpolated teachers are another approach. Their students typically do best at an intermediate $\lambda$, and long solutions give larger gradients and, from 3B students on, more concentrated update directions \citep{lan2026midistillation}, while the teacher changes its style until it stops thinking \citep{wu2026interpolation}. Behaviors also jump with the mixing ratio of expert assemblies \citep{klagges2025assembly}, and LoRA-distilled interpolated checkpoints are judged by accuracy alone \citep{saini2026thinkswitch}. These studies tune what the student learns from but leave open what its weights retain.

\subsection{Reading and Editing Model Internals}

LoRA updates can be compared as low-rank subspaces \citep{hu2021lora}. The reasoning part of a weight difference lies outside the instruct model's principal directions \citep[$S^3$;][]{ma2026spectral}, and task adapters overlap enough to stand in for one another, though without a chance baseline \citep{dieudonne2026adapter}.
Edited directions act as switches. Removing a refusal direction disables refusal \citep{arditi2024refusal}, removing about 2.5\% of an aligned model's ranks breaks its safety behavior \citep{wei2024brittleness}, and removing the top low-rank component of an LRM's change reduces overthinking \citep[Gold-Switch;][]{lee2025goldswitch}. Other edits act as dials on how long or how often a model reasons \citep{ma2025cotvalve,sun2025thinkedit,galichin2026covered,sheng2025planning,chen2025seal,jayabahu2026halt,venhoff2025steering}.
Sparse autoencoder (SAE) and transcoder features also locate reasoning behavior in distilled models. Features specific to a distilled model steer how long it thinks \citep{baek2025distilled}, and 2.4\% of the features of a distilled model's change are necessary and sufficient for its hesitation markers \citep{hu2026transcoder}. Distillation can implant new mechanisms \citep{venhoff2026base} or repurpose existing directions \citep{ward2025repurposes}. Yet many SAE reasoning features respond to a few cue tokens injected into non-reasoning text \citep{ma2026sae}.
These studies examine one model or model pair at a time \citep{lee2025goldswitch,ma2026spectral,hu2026transcoder,jiralerspong2026crossarch}. Unlike them, we vary the teacher continuously and edit a shared subspace that largely empties once the teacher stops thinking.

\section{Method}
\label{sec:method}

\subsection{Distillation from an Interpolated Teacher}
\label{sec:prelim}

An LRM with weights $\theta_R$ thinks before answering, while an instruct model $\theta_I$ fine-tuned from the same base answers briefly. The interpolated teacher with ratio $\lambda \in [0,1]$ has weights
\begin{equation}
\theta_T(\lambda) = (1-\lambda)\,\theta_R + \lambda\,\theta_I
\label{eq:teacher}
\end{equation}
\citep{wortsman2022soups,ilharco2023task}, so $\lambda=0$ is the LRM and $\lambda=1$ the instruct model. For each problem $x$ the teacher samples a solution $y$, and the correct, complete solutions form the training set $D_\lambda$. A student $\theta_S$ learns from the teacher by sequence-level distillation \citep{kim2016sequence},
\begin{equation}
\mathcal{L}(\theta_S) = -\,\mathbb{E}_{(x,y)\sim D_\lambda} \textstyle\sum_t \log p_{\theta_S}(y_t \mid x, y_{<t}),
\label{eq:kd}
\end{equation}
where $\theta_S$ is a LoRA adapter on a frozen student. We call a student's reflective markers, monologue format and long solutions its thinking-like behavior, and write thinking for short.

\subsection{Subspace Analysis}
\label{sec:subspace}

\paragraph{Overlap.}
An adapter's update directions are the top $r$ left singular vectors $U$ of each layer's update (Eq.~\eqref{eq:1}). This separates direction from magnitude, as polar-factored adapters do \citep{lion2025polar}. Adapters agree by the overlap $\phi$ (Eq.~\eqref{eq:2}) \citep{hu2021lora}, weighted over modules by $\lVert \Delta W \rVert_F$. It is 1 for identical subspaces and $r/d$ in expectation for random ones of output dimension $d$.
\begin{gather}
\Delta W = \tfrac{\alpha}{r} BA = U\Sigma Q^\top, \quad U = [\,u_1, \dots, u_r\,] \label{eq:1}\\
\phi(\Delta W_i, \Delta W_j) = \lVert U_i^\top U_j \rVert_F^2 / r \label{eq:2}
\end{gather}

\paragraph{Comparison.}
$R$ divides the overlap between adapters with different teachers by the overlap between adapters that differ only in data (Eq.~\eqref{eq:3}). Data halves $a$, $b$ and the seed enter both equally, so data effects cancel.
\begin{equation}
R(\lambda) = \frac{\phi(\Delta W_a^{0}, \Delta W_b^{\lambda}) + \phi(\Delta W_b^{0}, \Delta W_a^{\lambda})}{2\,\phi(\Delta W_a^{\lambda}, \Delta W_b^{\lambda})}
\label{eq:3}
\end{equation}
$R=1$ means a teacher effect at the level of the data difference, and $R$ near 0 a sharp change in the update directions.
We do not fix a switch $\lambda$ in advance. An exact permutation test checks whether drops in $R$ and in the teacher's think rate or markers correlate over $\lambda$ intervals (Appendix~\ref{app:details}).

\paragraph{Extraction.}
We take as the switch the first $\lambda$ interval in which the teacher's think rate drops by more than half. In all three pairs this is also its largest drop. We then average the projectors of the $n$ adapters before the switch into $P$. Our rule keeps the eigenvectors of $P$ whose eigenvalue exceeds 0.5 as the shared region $V$ (Eq.~\eqref{eq:4}). The eigenvalue is the mean squared share of the direction inside an adapter's update subspace.
\begin{equation}
P = \tfrac{1}{n}\textstyle\sum_{t} U_t U_t^\top,\;
V = \{v : Pv = \mu v,\, \mu > 0.5\}
\label{eq:4}
\end{equation}
An adapter's coverage of $V$ is $\lVert U_t^\top V \rVert_F^2 / k$ for $k = \dim V$, against chance $r/d$. Scores and edits use a $V$ built without that adapter (7.2--8.4 directions).

\subsection{Subspace Interventions}
\label{sec:interventions}

We remove the shared region, graft it into the $\lambda=1$ adapter, and amplify it. Controls edit other directions, and Appendix~\ref{app:details} gives the energy each takes.
\begin{gather}
\Delta W' = (I - VV^\top)\,\Delta W \label{eq:5}\\
\Delta W'(\lambda{=}1) = \Delta W(\lambda{=}1) + \beta\, VV^\top \Delta W(\lambda{=}0) \label{eq:6}\\
\Delta W' = \lVert \Delta W \rVert_F \, \dfrac{\Delta W + \gamma\, VV^\top \Delta W}{\lVert \Delta W + \gamma\, VV^\top \Delta W \rVert_F} \label{eq:7}
\end{gather}
We remove $V$ from the $\lambda=0$ adapter to test whether it is needed for thinking (Eq.~\eqref{eq:5}) \citep{arditi2024refusal}. We graft only its $\lambda=0$ component onto the $\lambda=1$ adapter, with $\beta=1$ and $2$, to test whether it is enough (Eq.~\eqref{eq:6}) \citep{panigrahi2023skill}. We also raise only its weight at a fixed norm, with $\gamma=1$ and $4$ or with the whole update moved into $V$ (Eq.~\eqref{eq:7}) \citep{ma2025cotvalve,galichin2026covered,sheng2025planning}. This saturation check can show a graded response above the threshold but cannot rule one out.
Removal acts on $B$, so the removal controls target the region's $\lVert B \rVert^2$ per module, taking it from the highest-energy directions outside $V$ or scaling the whole update down by it. The first reaches 0.46--0.98 of the region's $\lVert \Delta W \rVert^2$. Top and random removal take at least its $\lVert \Delta W \rVert^2$ from the largest singular components \citep{lee2025goldswitch} or random directions of the update's span. Grafting controls add the region's $\lVert \Delta W \rVert^2$ from outside $V$, or the whole $\lambda=0$ update at its $\lVert B \rVert^2$. We also edit the $\lambda=1$ adapter alike and cut the adapter at equal length \citep{muennighoff2025s1}.

\subsection{Experimental Setup}
\label{sec:setup}

We distill three teacher pairs into three students at six $\lambda$ with three seeds.\footnote{\url{https://github.com/kimtaeyeong1229/cot-distill-lora-subspace}} We measure thinking by think rate, reflective markers, monologue format and length, with SAE features as an auxiliary read-out.

\paragraph{Teachers.}
To narrow the cause to solution style, we vary only the teacher, at six values $\lambda \in \{0, 0.2, \dots, 1.0\}$ of Eq.~\eqref{eq:teacher}.
Pair A mixes R1-Distill-Qwen-1.5B and Qwen2.5-Math-1.5B-Instruct, pair B Qwen3-4B-Thinking-2507 and Qwen3-4B-Instruct-2507, and pair C R1-Distill-Llama-8B and Llama-3.1-8B-Instruct \citep{guo2025deepseekr1,yang2024qwen25math,yang2025qwen3,grattafiori2024llama3}. Pairs B and C are those of \citet{wu2026interpolation}, whose $\lambda$ runs the other way. Family and scale are confounded across pairs, so we make no claim about scale. All three pairs test where the adapters' update subspaces shift and what removing, grafting or amplifying the region does. Pair A, whose teacher keeps its think tag while solutions halve, is the only contrast separating thinking from length.

\paragraph{Students, data and training.}
The students are Qwen2.5-0.5B-Instruct, Llama-3.2-1B-Instruct, and OLMo-2-1B-Instruct \citep{qwen2025qwen25,meta2024llama32,olmo2025olmo2}. At these sizes students learn better from short solutions \citep{li2025small}. They come from three families so that no result is specific to one family or its contamination \citep{wu2026memorization}. Each pair-student combination is a cell.
All teachers solve the same 4,000 MATH training problems \citep{hendrycks2021math}. We train only on problems that the teacher at every $\lambda$ solves correctly and completely (2,113, 2,261 and 1,015 for pairs A, B and C). The $\lambda=1$ teacher thus solves every training problem with a short answer. Every $\lambda=0$ solution is a complete long solution these problems did not need, the overthinking of \citet{chen2025overthinking}.
Each student gets a LoRA adapter ($r=16$, $\alpha=32$), used for measurement, not efficiency, since it stores the update in two small matrices comparable across runs. For each cell and $\lambda$, we train three seeds on the full set. To isolate the data effect, we also train on two disjoint halves (Appendix~\ref{app:measure}).

\paragraph{Evaluation.}
Each seed's adapter solves the first 200 MATH-500 problems \citep{lightman2023verify} twice. We report seed means $\pm$ standard deviations, with pass@1 accuracy over all and over completed responses.
For teachers, thinking is the think rate, the share of responses with \texttt{</think>} \citep{wu2026interpolation}. For students and teacher solutions, it is the rate of reflective markers (\emph{wait}, \emph{recheck}, etc.) \citep{yeo2025demystifying}.
Keywords are only a surface proxy for long-solution reasoning \citep{chen2026molecular}, so we also report tokens and the generation-limit hit rate \citep{goyal2026cap}. The Llama student is also read with a public SAE. We keep features that separate $\lambda=0$ from $\lambda=1$ teacher solutions and survive a marker-insertion check adapted from \citet{ma2026sae} and our own marker-deletion check. We call these the SAE thinking features.
We also test GSM8K \citep{cobbe2021training}, GPQA-Diamond \citep{rein2023gpqa} and untrained students (Appendix~\ref{app:details}).

\section{Where the Adapter Switches}\label{sec:switch}

\begin{figure*}[t]
\centering
\includegraphics[width=0.78\textwidth]{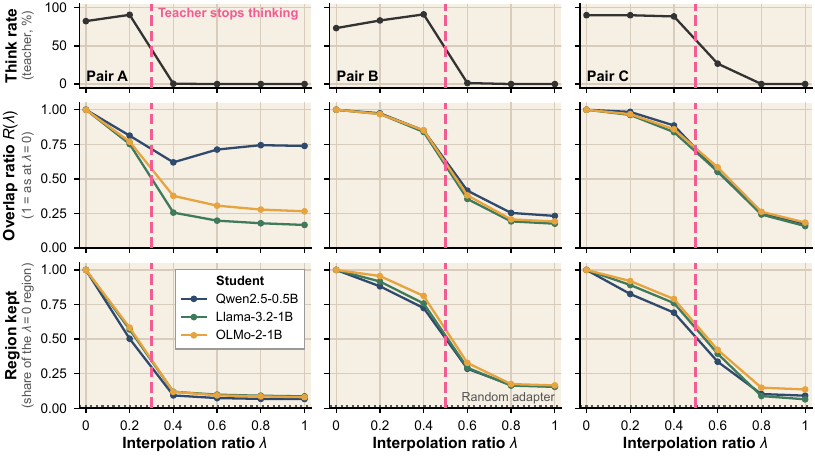}
\caption{Panels show teacher think rate (top), overlap ratio $R(\lambda)$ (middle) and share of the $\lambda=0$ region kept (bottom) per student (seed means), with dotted lines for a random adapter and dashed lines for the switch (\S\ref{sec:subspace}). $R$ and the kept share are 1 at $\lambda=0$, and A/Qwen's $R$ recovers after the switch as its data-half overlap falls (Appendix~\ref{app:measure}). The bottom panel divides coverage of the region built from the three $\lambda=0$ adapters by that of half-data $\lambda=0$ adapters, a post hoc reference (the pre-specified leave-$\lambda$-out coverage is in Table~\ref{tab:length}).}
\label{fig:switch}
\end{figure*}

\subsection{Student Outputs}\label{sec:outputs}

Student outputs follow the teacher, with thinking and length falling together, so outputs alone cannot separate them. As each teacher drops \texttt{</think>} and its markers fall, its students' markers track this change closely in every cell. Meanwhile accuracy rises, mostly from recovered truncation, but falls over completed responses (fewer, easier problems at $\lambda=0$) in eight of nine cells (Table~\ref{tab:2}). Students score below their untrained models at 41 of 54 cell--$\lambda$ points, including $\lambda=1$ in six of nine cells (Tables~\ref{tab:2} and~\ref{tab:untrained}). No edit raises the mean above that level, so we analyze what students store, not accuracy.

\subsection{Adapter Directions}\label{sec:directions}

In each pair, the adapters' update subspaces shift most around where their own teacher stops thinking, which is also where training solutions shorten most. $R$ drops most at $\lambda = 0.2 \to 0.4$ in pair A (narrowly in A/Qwen), at $0.4 \to 0.6$ in pair B and over $0.4 \to 0.6 \to 0.8$ in pair C (Figure~\ref{fig:switch}, middle), where the teacher also changes, so the adapter follows its own teacher, not $\lambda$.
Outside A/Qwen, interval drops closely track the teacher's. The exact test pooling the pairs gives $p = 0.0035$, while no single pair reaches $p<0.05$ on the think rate (Table~\ref{tab:4}).
Training-length drops give $p = 0.0002$, so the test only locates the change.

\subsection{Directions Shared Before the Switch}\label{sec:shared}

Before the switch, adapters share a region of 5--8 (weighted) of their 16 directions. In pair A it keeps about half while training solutions halve, and it largely empties after thinking stops. The pre-specified leave-$\lambda$-out coverage (Table~\ref{tab:length}) is symmetric by construction over pair A's held interval ($\lambda = 0 \to 0.2$, where the teacher keeps thinking), so it cannot show retention there. In pairs B and C, a region built from $\lambda \le 0.2$ alone is covered 0.48--0.55 by the $\lambda=0.4$ adapters and 0.19--0.30 by the $\lambda=0.6$ adapters. Pair A's $\lambda=0.4$ adapters keep 0.09--0.12 of the $\lambda=0$ region (Figure~\ref{fig:switch}, bottom). The pre-switch adapters cover 41--57\% of this region, against 5--19\% for post-switch adapters and 0.5--1.8\% for a random adapter (Table~\ref{tab:5}). However, other seeds at a pre-switch adapter's own $\lambda$ help build the region it is scored against.

\begin{table}[t]
\centering
\footnotesize
\setlength{\tabcolsep}{4pt}
\renewcommand{\arraystretch}{0.92}
\begin{adjustbox}{max width=\columnwidth}\begin{tabular}{lllrrrrr}
\toprule
Pair & $\lambda$ & Class & $\Delta\log L$ & $\Delta$Think & Markers & Kept & Loss \\
 & & & & (pp) & ($\times$) & & \\
\midrule
A & 0--0.2 & Held & 0.67 & +8 & 0.32 & 0.55 & 0.89 \\
A & 0.2--0.4 & Switch & 0.80 & $-$90 & 0.20 & 0.17 & 2.27 \\
\midrule
B & 0--0.2 & Held & 0.21 & +10 & 0.92 & 0.92 & 0.42 \\
B & 0.2--0.4 & Held & 0.19 & +8 & 0.89 & 0.87 & 0.73 \\
B & 0.4--0.6 & Switch & 0.68 & $-$90 & 0.46 & 0.42 & 1.28 \\
B & 0.6--0.8 & After & 0.51 & $-$1 & 0.85 & 0.47 & 1.49 \\
\midrule
C & 0--0.2 & Held & 0.22 & 0 & 0.89 & 0.88 & 0.59 \\
C & 0.2--0.4 & Held & 0.20 & $-$2 & 0.89 & 0.85 & 0.85 \\
C & 0.4--0.6 & Switch & 0.60 & $-$62 & 0.44 & 0.51 & 1.14 \\
C & 0.6--0.8 & Switch & 0.69 & $-$27 & 0.11 & 0.19 & 2.50 \\
\bottomrule
\end{tabular}
\end{adjustbox}
\caption{Teacher intervals. Columns give the log training-length drop, the change in think rate (pp) and the end-to-start ratio of the teacher's markers. Kept and Loss are student means (per student in Table~\ref{tab:length}) of the share of the region built at the interval's start kept relative to half-data adapters, and of its loss per unit of log-length drop. Pair A's switch/held loss ratio is 2.53--2.91 with a cross-seed build and 1.93--3.07 with two floors (Appendix~\ref{app:measure}), all post hoc like the reference. These ranges span students, seeds and builds and are not intervals.}
\label{tab:lenbody}
\end{table}

Because training solutions also shorten with $\lambda$, most at the switch, we also score the share of the region built at an interval's start that its end adapters keep, relative to half-data adapters (Table~\ref{tab:length}). This reference was chosen after the edit results were seen.
In pair A, solutions halve over the held interval and shorten similarly at the switch. Yet the switch loses about 2.5 times as much of the region per unit of log-length drop (Table~\ref{tab:lenbody}, Appendix~\ref{app:measure}). This is a single teacher contrast, in which the tag holds and the teacher's markers fall threefold. A pooled switch-indicator regression has an interval spanning zero, since pairs B and C confound length with class (Appendix~\ref{app:measure}). Over this interval the teacher keeps its think tag, recap, monologue opener and verification rate, while hesitation and attempts fall more than threefold (Table~\ref{tab:structure}). The region thus keeps about half while format holds and exploration falls, and largely empties after the format goes.
In pairs B and C, held intervals keep nearly every measure and switches shorten solutions about three times more (Table~\ref{tab:lenbody}). In most cells the edits switch format and hesitation together, so the edits alone cannot tell which the region carries. Same-seed adapters share their LoRA initialization, which raises their overlap, but the cross-seed build keeps the pair-A contrast (Table~\ref{tab:lenbody}). The whole update's cross-seed overlap with the $\lambda=0$ adapters also keeps about half over pair A's held interval and loses 1.5--2.0 times as much per unit at the switch, so only the region's excess at the switch is specific to it. Relative to the whole update, pair A's region loses no more over its held interval and a further 35--62\% at its switch. Pairs B and C lose at most a quarter at their first switch step. Pair C's region loses a further 36--64\% at its second switch step, where its teacher stops thinking altogether. Outside pair A the emptying thus partly tracks the whole update leaving the $\lambda=0$ directions.

\section{What the Shared Region Carries}\label{sec:carries}

\begin{figure*}[t]
\centering
\includegraphics[width=0.80\textwidth]{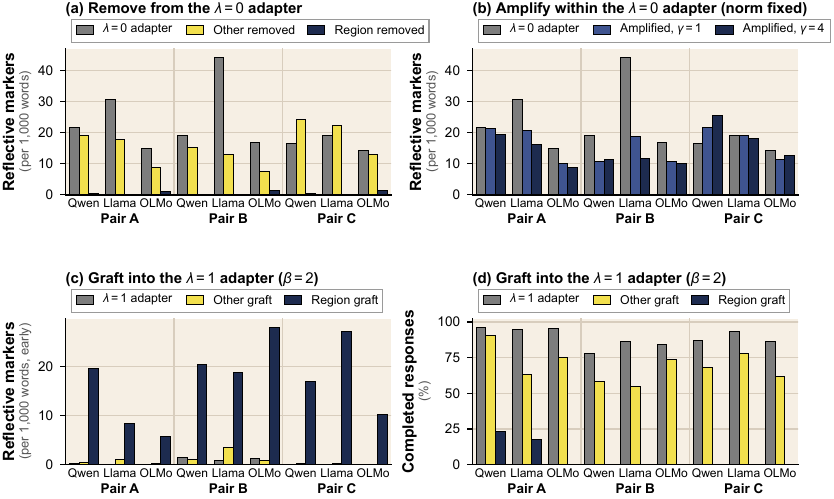}
\caption{Bars give seed means per cell, with the unedited adapter in gray, region edits in navy and the same edits outside it in yellow. Panels show whole-response markers of the $\lambda=0$ adapter after (a) removal and (b) amplification with $\gamma=1$ (blue) and $\gamma=4$ (navy), and (c) early-response markers, a metric chosen after the grafts looped (whole-response in Table~\ref{tab:9}), and (d) completed responses after grafting at $\beta=2$. Table~\ref{tab:editbody} adds top and random removal and the whole-update graft.}
\label{fig:states}
\end{figure*}

\begin{table}[t]
\centering
\footnotesize
\setlength{\tabcolsep}{4pt}
\renewcommand{\arraystretch}{0.88}
\begin{adjustbox}{max width=\columnwidth}\begin{tabular}{lrrr}
\toprule
 & Pair A & Pair B & Pair C \\
\midrule
\rowcolor{band}\multicolumn{4}{l}{\emph{Markers kept after removal (\%)}} \\
Region & 3.3 & 3.2 & 4.2 \\
Other directions & 68 & 51 & 118 \\
Top components & 28 & 6 & 11 \\
Random directions & 29 & 4 & 24 \\
\midrule
\rowcolor{band}\multicolumn{4}{l}{\emph{Accuracy gain over the length-matched cut (pp)}} \\
Region removed & 3.8 & 11.7 & 10.5 \\
\midrule
\rowcolor{band}\multicolumn{4}{l}{\emph{Early-marker gap recovered by grafting (\%)}} \\
Region, $\beta=1$ & 5 & 78 & 4 \\
Region, $\beta=2$ & 106 & 148 & 162 \\
Whole update, $\beta=2$ & 97 & 314 & 85 \\
\midrule
\rowcolor{band}\multicolumn{4}{l}{\emph{Whole-response gap recovered by grafting (\%)}} \\
Region, $\beta=2$ & 27 & 84 & 132 \\
Whole update, $\beta=2$ & 68 & 188 & 87 \\
\midrule
\rowcolor{band}\multicolumn{4}{l}{\emph{Completed responses (\%)}} \\
$\lambda=1$ host & 95 & 83 & 89 \\
Region graft, $\beta=2$ & 13.5 & 0.1 & 0.1 \\
\bottomrule
\end{tabular}
\end{adjustbox}
\caption{Edit effects as means over each pair's three students (per cell in Tables~\ref{tab:6}, \ref{tab:graftstats} and~\ref{tab:9}). Early markers are counted in the first 300 words, a metric chosen after the grafts looped. Whole-response markers were pre-specified. Recovery above 100\% is an overshoot, and the whole update is the $\lambda=0$ update at the region's $\lVert B \rVert^2$. Random directions lie in the update's span.}
\label{tab:editbody}
\end{table}

\subsection{Removing}\label{sec:remove}

Removing the shared region largely turns thinking off and leaves short complete solutions, which removing other directions does not. Removal from the $\lambda=0$ adapter drops whole-response reflective markers to 0.2--10\% of their level and early markers from 10--16 to 0.07--1.3 per 1,000 words. Removing other directions with at most the region's energy, or scaling down the whole update, leaves far more (Figure~\ref{fig:states}a, Tables~\ref{tab:editbody} and~\ref{tab:6}). In B/OLMo and C/OLMo that control removes only about half of the region's $\lVert \Delta W \rVert^2$ (Table~\ref{tab:energy}), and thinking survives it (Table~\ref{tab:6}).
In pairs B and C, energy-matched top removal \citep{lee2025goldswitch} keeps within 5 points of the region's marker share in four of the six cells, and B/Qwen is at the margin (6.0\% against 0.8\%). Its directions lie mostly inside the region (Appendix~\ref{app:details}). Specificity against the update's dominant directions is therefore shown only in pair A and C/Qwen (Tables~\ref{tab:editbody} and~\ref{tab:6}). Random directions of the update's span at the same $\lVert \Delta W \rVert^2$, which also take part of the region, keep 15--41\% of the markers in pair A and 28--31\% in C/Qwen and C/Llama but 1.6--14\% in pair B and C/OLMo. A same-size post-switch region removes just 0.06--0.33 of the region's $\lVert \Delta W \rVert^2$ yet leaves 9--23\% of the markers in B/Qwen, B/Llama, C/Qwen and C/Llama, while in pair A, B/OLMo and C/OLMo it raises them to 104--181\% (Table~\ref{tab:6}).
Removal gives the highest mean accuracy among single-adapter edits of the $\lambda=0$ adapter (Tables~\ref{tab:general} and~\ref{tab:7}). It beats the length-matched cut in all nine cells in point estimate, with the interval above zero in seven. The mean gain is +8.6 pp, with a problem-bootstrap interval of [6.5, 10.8] when cells are fixed. Treating cells or teachers as units widens the t-interval to [3.3, 14.0] or [$-2.0$, 19.3]. At these units its advantage over top removal spans zero, and it reverses in A/OLMo and C/OLMo (Table~\ref{tab:stats}, Appendix~\ref{app:edits}).
Removal takes the structure measures to the untrained level, with two exceptions. Verification survives in A/OLMo and C/OLMo (59\% and 90\% of the excess kept), and the monologue opener in A/Qwen and A/Llama (96\% and 26\% of responses, 87\% and 25\% with ``Okay''). Other-direction removal keeps most of these excesses (Table~\ref{tab:studstruct}). A/Qwen also keeps part of its monologue sentences (0.17 against 0.45). A/Qwen and A/Llama write 1.5--1.9 times the untrained length, and A/Qwen scores 14.8 pp below untrained accuracy (Table~\ref{tab:untrained}). In pairs B and C the removed student comes within 1.3 pp of untrained accuracy. Its Cohen's $\kappa$ on answer correctness with the untrained student is 0.37--0.54, against 0.35--0.44 for the $\lambda=1$ adapter and an untrained self-ceiling of 0.46--0.57 (Table~\ref{tab:untrained}). There the region cannot be told from the whole update. The region is needed for thinking in all nine cells (three teachers), but specific beyond the dominant directions only in pair A and C/Qwen.

\subsection{Grafting}\label{sec:graft}

Grafted into the $\lambda=1$ adapter at twice its weight, the shared region turns early reflective markers on, as the whole update does, but the student rarely finishes. At its own weight the region barely moves markers in pairs A and C (two usable doses). Grafted students loop, so we score grafts on early markers in the first 300 words, a metric added after seeing the grafts. At $\beta=1$ the region recovers only 2--8\% of the early-marker gap in pairs A and C, though responses already lengthen.
At $\beta=2$, which adds 4$\times$ the energy of $\beta=1$, the region turns early markers on in all nine cells. It recovers 56--218\% of the early-marker gap in pairs A and C (Table~\ref{tab:graftstats}), while pair B is already partly on at $\beta=1$. The graft restores hesitation beyond the $\lambda=0$ level in the nine-cell mean, and the monologue opener. It restores verification only unevenly (Appendix~\ref{app:edits}) and the recap only where grafts finish (A/Qwen and A/Llama, Table~\ref{tab:studstruct}). A logit lens does not separate the region from the post-switch region on its pre-specified score (Appendix~\ref{app:edits}).
The whole $\lambda=0$ update grafted at the same energy also turns early markers on, in eight of nine cells (all but C/Qwen). It recovers more of the pre-specified gap than the region in seven cells, finishes more often in seven (one tie) and is more accurate in all nine. The region leads on the post hoc early metric in five cells and on whole-response markers only in B/Qwen and C/Qwen (Tables~\ref{tab:editbody} and~\ref{tab:9}). On the pre-specified metric, the graft thus separates the region from other directions in all cells but B/Llama. It does not separate the region from the update's dominant directions.
The $\lambda=1$ host finishes 78--96\% of responses. Completion stays low when the graft carries the region, at 0--23\% for the region and about as low for the whole update, which puts 0.34--0.59 of its $\lVert B \rVert^2$ in the region (Table~\ref{tab:energy}). It stays high with other directions (Figure~\ref{fig:states}d, Table~\ref{tab:9}). A/Qwen and A/Llama grafts finish about as often as their $\lambda=0$ donors (Tables~\ref{tab:2} and~\ref{tab:9}). OLMo grafts almost never finish, confounded by its truncated training data and 3,840-token cap. Most region grafts loop (69--96\%, Table~\ref{tab:loop}), and larger $\beta$ degenerates (Appendix~\ref{app:edits}).

\subsection{Amplifying}\label{sec:amplify}

At a fixed norm the region's weight can rise only 1.2--1.6$\times$. This moves early markers 0.87--1.79$\times$ and often lowers whole-response markers, and the region may act as a dial in two pair-C cells. Whole-response markers fall 41--74\% at $\gamma=4$ in five of nine cells and rise only in C/Qwen (Figure~\ref{fig:states}b, Table~\ref{tab:10}).
In C/Qwen and C/Llama early markers rise faster than the region's weight. They rise about proportionally in A/Qwen and C/OLMo and more slowly elsewhere.

\subsection{Student Features}\label{sec:sae}

The Llama student's SAE thinking features turn off and on with removal and grafting in pairs A and C. Each edited model reads its own responses, so the three Llama cells' features mix text and edit and may reflect style.
The features track markers at $r = 0.88$ in pair A and $0.92$ in C (0.86 and 0.91 with markers deleted). Pair B gives $r = -0.08$, and each $r$ pools 57 condition and seed points, mostly edited against unedited adapters (Table~\ref{tab:11}).
Removal cuts the features far more than removing other directions. Grafting at $\beta=2$ restores them to 51\% and 69\% of the $\lambda=0$ level in pairs A and C (46\% and 65\% of the gap from $\lambda=1$). In pair B the graft leaves the whole-response features at the $\lambda=1$ level, though over the first 400 tokens they rise from 0.17 to 0.45.

\section{How General the Switch Is}\label{sec:general}

\subsection{Other Benchmarks}\label{sec:ood}

Removing the MATH-trained region largely turns thinking off on GSM8K and GPQA-Diamond too. Removal leaves at most 10.4\% of the whole-response markers on both benchmarks, far fewer than removing other directions (Figure~\ref{fig:ood}).
GSM8K accuracy rises (Table~\ref{tab:general}), partly from resolved truncation. On GPQA-Diamond every condition is below chance, so we compare only markers and length there.

\begin{table}[t]
\centering
\footnotesize
\setlength{\tabcolsep}{4pt}
\begin{tabular}{lrrrr}
\toprule
 & \multicolumn{2}{c}{GSM8K} & \multicolumn{2}{c}{GPQA-Diamond} \\
\cmidrule(lr){2-3}\cmidrule(lr){4-5}
Condition & Tokens & Acc. & Tokens & Acc. \\
\midrule
$\lambda=0$ & 3,596 & 34.2 & 3,319 & 13.7 \\
Region removed & 353 & \underline{47.6} & 938 & 20.1 \\
Other removed & 3,268 & 39.3 & 2,687 & 17.6 \\
$\lambda=1$ & 411 & \textbf{48.5} & 1,226 & 22.0 \\
\bottomrule
\end{tabular}

\vspace{6pt}
\begin{adjustbox}{max width=\columnwidth}\begin{tabular}{lrrr}
\toprule
Condition & Markers & Tokens & Acc. \\
\midrule
$\lambda=0$ & 22.0 & 4,820 & 13.9 \\
Region removed & 0.6 & 1,075 & 24.7 \\
Two-seed region removed & 1.7 & 1,403 & 21.7 \\
Gold-Switch-style removal ($\tau$ rank) & 5.2 & 2,585 & 19.1 \\
Length-matched cut & 15.0 & 1,076 & 16.0 \\
Think-skip (adapted) & 16.2 & 4,570 & 13.9 \\
Other removed (two-seed region) & 18.2 & 4,616 & 16.1 \\
\midrule
Merged with $\lambda=1$ & 3.5 & 1,429 & \underline{25.4} \\
$\lambda=1$ & 1.6 & 1,218 & 24.3 \\
Untrained & 0.4 & 813 & \textbf{26.9} \\
\bottomrule
\end{tabular}\end{adjustbox}

\caption{Nine-cell means. The top block gives tokens and accuracy (\%) of the $\lambda=0$ adapter on GSM8K and GPQA-Diamond. The bottom block gives markers per 1,000 words, tokens and accuracy of ways to turn thinking off on MATH-500, and rows below the rule need a second adapter or none. Untrained students ran on MATH-500 only. Bold and underline mark the two best GSM8K and MATH-500 accuracies.}
\label{tab:general}
\end{table}

\subsection{A Region from Two $\lambda=0$ Seeds}\label{sec:noninterp}

We build a region from two $\lambda=0$ adapters that differ only in seed and remove it from a third without retraining. It spans more than half of the shared region (Table~\ref{tab:cuts}), and its removal leaves far fewer markers than removing other directions. Since $\lambda=0$ adapters also help build the shared region, this overlap is not independent evidence, though these two seeds share no initialization with the edited adapter. Other ways to turn thinking off (Table~\ref{tab:general}) leave more markers or need a second adapter \citep{lee2025goldswitch,ma2025nothinking,wu2025merging}.

\section{Conclusion}\label{sec:conclusion}

In small LoRA students distilled from interpolated teachers, we identified a shared adapter subspace tied to thinking-like behavior. The adapters' update subspaces shift where the teacher stops thinking, and the subspace shared before then largely empties. Removing it largely turns this behavior off in all nine cells and on other benchmarks, and grafting it at twice its weight turns early markers on, though the student rarely finishes. Its specificity beyond the dominant directions holds only in pair A and C/Qwen, and its separation from length only in pair A, so separating thinking from solution length and scaling up are future work.

\clearpage
\section*{Limitations}
\phantomsection\label{sec:limitations}

We study off-policy distillation on MATH into students of at most 1B parameters, with LoRA as an instrument rather than the standard way to distill. Full fine-tuning updates differ in structure \citep{shuttleworth2025lora}, so the region may not arise there. Model family and scale change together across the pairs, and our only evidence outside math is GPQA-Diamond. Students score below their untrained models at 41 of 54 cell--$\lambda$ points, no $\lambda$ beats it in five cells, and accuracy peaks after the switch ($\lambda=0.8$ or 1) in all nine. At $\lambda=0$ they write 2.8--7.0k tokens from 2.1--3.1k-token training solutions and loop in 14--80\% of responses. They may sit in the dip that \citet{luo2025valley} report, which more data fully recovers only for larger models, so we study what the students store, not a better student.

In pairs B and C, removing the region returns the student to untrained behavior. There we cannot separate ``the region carries thinking'' from ``the region carries most of what the teacher added''. Removing random directions of the update's own span at the same energy also leaves at most 14\% of the markers in pair B and C/OLMo, against 28--31\% in C/Qwen and C/Llama. A same-size post-switch region also cuts markers in four B and C cells, and grafting the whole $\lambda=0$ update at $\beta=2$ also turns markers on in eight of nine cells. Thinking and training length come apart only in pair A's first interval, where the tag holds but the teacher's markers fall. Pair B's 0.6 to 0.8 interval loses more of its post-switch region per unit of length than its switch (Appendix~\ref{app:measure}). Our filter favors easier problems (level-5 share 30\% to 16--17\%), and three seeds per cell give intervals with little seed variation.

We measure thinking by markers and, for Llama only, by SAE features that each edited model reads from its own text and that may reflect style. At a fixed norm the region's weight rises only 1.2--1.6$\times$, so above that range the region may act as a dial in C/Qwen and C/Llama. Grafting has two usable doses and no cross-seed graft. Graft completion is measured against the $\lambda=1$ host, the $\lambda=0$ donor finishes about as rarely in A/Qwen and A/Llama, and OLMo's grafts are confounded by its 3,840-token cap and truncated training solutions.

\section*{Use of AI Assistants}
AI assistants (Claude via Claude Code, and Gemini for early drafts of Figure~\ref{fig:overview}) were used to write and run code, search the literature, suggest analyses, and draft and revise text. The authors verified all generated content and take full responsibility for the paper.

\bibliography{custom}

\begin{thebibliography}{63}
\providecommand{\natexlab}[1]{#1}

\bibitem[{Arditi et~al.(2024)Arditi, Obeso, Syed, Paleka, Panickssery, Gurnee,
  and Nanda}]{arditi2024refusal}
Andy Arditi, Oscar Obeso, Aaquib Syed, Daniel Paleka, Nina Panickssery, Wes
  Gurnee, and Neel Nanda. 2024.
\newblock \href {https://doi.org/10.52202/079017-4322} {Refusal in language
  models is mediated by a single direction}.
\newblock In \emph{Advances in Neural Information Processing Systems},
  volume~37, pages 136037--136083. Curran Associates, Inc.

\bibitem[{Baek and Tegmark(2025)}]{baek2025distilled}
David~D. Baek and Max Tegmark. 2025.
\newblock \href {https://arxiv.org/abs/2503.03730} {Towards understanding
  distilled reasoning models: A representational approach}.
\newblock \emph{arXiv preprint arXiv:2503.03730}.

\bibitem[{Boizard et~al.(2026)Boizard, Gisserot-Boukhlef, Haddad, Hudelot, and
  Colombo}]{boizard2026scale}
Nicolas Boizard, Hippolyte Gisserot-Boukhlef, Kevin~El Haddad, C{\'e}line
  Hudelot, and Pierre Colombo. 2026.
\newblock \href {https://arxiv.org/abs/2509.22193} {Scale or reason? {A}
  compute-equivalent analysis of reasoning distillation}.
\newblock \emph{arXiv preprint arXiv:2509.22193}.

\bibitem[{Chen et~al.(2026)Chen, Du, Li, Liu, Duan, Guo, Liu, Liu, Yang, Zhang,
  Qin, Che, and Huang}]{chen2026molecular}
Qiguang Chen, Yantao Du, Ziniu Li, Jinhao Liu, Songyao Duan, Jiarui Guo,
  Minghao Liu, Jiaheng Liu, Tong Yang, Ge~Zhang, Libo Qin, Wanxiang Che, and
  Wenhao Huang. 2026.
\newblock \href {https://arxiv.org/abs/2601.06002} {The molecular structure of
  thought: Mapping the topology of long chain-of-thought reasoning}.
\newblock \emph{arXiv preprint arXiv:2601.06002}.

\bibitem[{Chen et~al.(2025{\natexlab{a}})Chen, Zhang, Hong, Kundu, and
  Wang}]{chen2025seal}
Runjin Chen, Zhenyu Zhang, Junyuan Hong, Souvik Kundu, and Zhangyang Wang.
  2025{\natexlab{a}}.
\newblock \href {https://openreview.net/forum?id=klPszYDIRT} {{SEAL}: Steerable
  reasoning calibration of large language models for free}.
\newblock In \emph{Second Conference on Language Modeling}.

\bibitem[{Chen et~al.(2025{\natexlab{b}})Chen, Xu, Liang, He, Pang, Yu, Song,
  Liu, Zhou, Zhang, Wang, Tu, Mi, and Yu}]{chen2025overthinking}
Xingyu Chen, Jiahao Xu, Tian Liang, Zhiwei He, Jianhui Pang, Dian Yu, Linfeng
  Song, Qiuzhi Liu, Mengfei Zhou, Zhuosheng Zhang, Rui Wang, Zhaopeng Tu,
  Haitao Mi, and Dong Yu. 2025{\natexlab{b}}.
\newblock \href {https://proceedings.mlr.press/v267/chen25bx.html} {Do {NOT}
  think that much for 2+3=? {O}n the overthinking of long reasoning models}.
\newblock In \emph{Proceedings of the 42nd International Conference on Machine
  Learning}, volume 267 of \emph{Proceedings of Machine Learning Research},
  pages 9487--9499. PMLR.

\bibitem[{Cobbe et~al.(2021)Cobbe, Kosaraju, Bavarian, Chen, Jun, Kaiser,
  Plappert, Tworek, Hilton, Nakano, Hesse, and Schulman}]{cobbe2021training}
Karl Cobbe, Vineet Kosaraju, Mohammad Bavarian, Mark Chen, Heewoo Jun, Lukasz
  Kaiser, Matthias Plappert, Jerry Tworek, Jacob Hilton, Reiichiro Nakano,
  Christopher Hesse, and John Schulman. 2021.
\newblock \href {https://arxiv.org/abs/2110.14168} {Training verifiers to solve
  math word problems}.
\newblock \emph{arXiv preprint arXiv:2110.14168}.

\bibitem[{Dieudonn{\'e} et~al.(2026)Dieudonn{\'e}, Lanzillotta, Simsar,
  Barinka, and Hofmann}]{dieudonne2026adapter}
Tanguy Dieudonn{\'e}, Giulia Lanzillotta, Enis Simsar, Louis Barinka, and
  Thomas Hofmann. 2026.
\newblock \href {https://arxiv.org/abs/2606.28117} {When one adapter speaks for
  many: Discovering low-rank redundancy in continual fine-tuning}.
\newblock \emph{arXiv preprint arXiv:2606.28117}.

\bibitem[{Galichin et~al.(2026)Galichin, Dontsov, Druzhinina, Razzhigaev,
  Rogov, Tutubalina, and Oseledets}]{galichin2026covered}
Andrey Galichin, Alexey Dontsov, Polina Druzhinina, Anton Razzhigaev, Oleg~Y.
  Rogov, Elena Tutubalina, and Ivan Oseledets. 2026.
\newblock \href {https://doi.org/10.1609/aaai.v40i36.40334} {I have covered all
  the bases here: Interpreting reasoning features in large language models via
  sparse autoencoders}.
\newblock \emph{Proceedings of the AAAI Conference on Artificial Intelligence},
  40(36):30771--30779.

\bibitem[{Goyal and Ray(2026)}]{goyal2026cap}
Ankit Goyal and Jaideep Ray. 2026.
\newblock \href {https://arxiv.org/abs/2608.04160} {Mind the cap: Output-budget
  regimes change the measured multilingual reasoning gap}.
\newblock \emph{arXiv preprint arXiv:2608.04160}.

\bibitem[{Grattafiori et~al.(2024)Grattafiori, Dubey, Jauhri, Pandey, Kadian,
  Al-Dahle, Letman, Mathur, Schelten, Vaughan, Yang, Fan, Goyal, Hartshorn,
  Yang, Mitra, Sravankumar, Korenev, Hinsvark, Rao, Zhang, Rodriguez,
  Gregerson, Spataru, Roziere, Biron, Tang, Chern, Caucheteux, Nayak, Bi,
  Marra, McConnell, Keller, Touret, Wu, Wong, Ferrer, Nikolaidis, Allonsius,
  Song, Pintz, Livshits, Wyatt, Esiobu, Choudhary, Mahajan, Garcia-Olano,
  Perino, Hupkes, Lakomkin, AlBadawy, Lobanova, Dinan, Smith, Radenovic,
  Guzm{\'a}n, Zhang, Synnaeve, Lee, Anderson, Thattai, Nail, Mialon, Pang,
  Cucurell, Nguyen, Korevaar, Xu, Touvron, Zarov, Ibarra, Kloumann, Misra,
  Evtimov, Zhang, Copet, Lee, Geffert, Vranes, Park, Mahadeokar, Shah, van~der
  Linde, Billock, Hong, Lee, Fu, Chi, Huang, Liu, Wang, Yu, Bitton, Spisak,
  Park, Rocca, Johnstun, Saxe, Jia, Alwala, Prasad, Upasani, Plawiak, Li,
  Heafield, Stone, El-Arini, Iyer, Malik, Chiu, Bhalla, Lakhotia,
  Rantala-Yeary, van~der Maaten, Chen, Tan, Jenkins, Martin, Madaan, Malo,
  Blecher, Landzaat, de~Oliveira, Muzzi, Pasupuleti, Singh, Paluri, Kardas,
  Tsimpoukelli, Oldham, Rita, Pavlova, Kambadur, Lewis, Si, Singh, Hassan,
  Goyal, Torabi, Bashlykov, Bogoychev, Chatterji, Zhang, Duchenne,
  {\c{C}}elebi, Alrassy, Zhang, Li, Vasic, Weng, Bhargava, Dubal, Krishnan,
  Koura, Xu, He, Dong, Srinivasan, Ganapathy, Calderer, Cabral, Stojnic,
  Raileanu, Maheswari, Girdhar, Patel, Sauvestre, Polidoro, Sumbaly, Taylor,
  Silva, Hou, Wang, Hosseini, Chennabasappa, Singh, Bell, Kim, Edunov, Nie,
  Narang, Raparthy, Shen, Wan, Bhosale, Zhang, Vandenhende, Batra, Whitman,
  Sootla, Collot, Gururangan, Borodinsky, Herman, Fowler, Sheasha, Georgiou,
  Scialom, Speckbacher, Mihaylov, Xiao, Karn, Goswami, Gupta, Ramanathan,
  Kerkez, Gonguet, Do, Vogeti, Albiero, Petrovic, Chu, Xiong, Fu, Meers,
  Martinet, Wang, Wang, Tan, Xia, Xie, Jia, Wang, Goldschlag, Gaur, Babaei,
  Wen, Song, Zhang, Li, Mao, Coudert, Yan, Chen, Papakipos, Singh, Srivastava,
  Jain, Kelsey, Shajnfeld, Gangidi, Victoria, Goldstand, Menon, Sharma,
  Boesenberg, Baevski, Feinstein, Kallet, Sangani, Teo, Yunus, Lupu, Alvarado,
  Caples, Gu, Ho, Poulton, Ryan, Ramchandani, Dong, Franco, Goyal, Saraf,
  Chowdhury, Gabriel, Bharambe, Eisenman, Yazdan, James, Maurer, Leonhardi,
  Huang, Loyd, De~Paola, Paranjape, Liu, Wu, Ni, Hancock, Wasti, Spence,
  Stojkovic, Gamido, Montalvo, Parker, Burton, Mejia, Liu, Wang, Kim, Zhou, Hu,
  Chu, Cai, Tindal, Feichtenhofer, Gao, Civin, Beaty, Kreymer, Li, Adkins, Xu,
  Testuggine, David, Parikh, Liskovich, Foss, Wang, Le, Holland, Dowling,
  Jamil, Montgomery, Presani, Hahn, Wood, Le, Brinkman, Arcaute, Dunbar,
  Smothers, Sun, Kreuk, Tian, Kokkinos, Ozgenel, Caggioni, Kanayet, Seide,
  Florez, Schwarz, Badeer, Swee, Halpern, Herman, Sizov, Zhang,
  Lakshminarayanan, Inan, Shojanazeri, Zou, Wang, Zha, Habeeb, Rudolph, Suk,
  Aspegren, Goldman, Zhan, Damlaj, Molybog, Tufanov, Leontiadis, Veliche, Gat,
  Weissman, Geboski, Kohli, Lam, Asher, Gaya, Marcus, Tang, Chan, Zhen,
  Reizenstein, Teboul, Zhong, Jin, Yang, Cummings, Carvill, Shepard, McPhie,
  Torres, Ginsburg, Wang, Wu, U, Saxena, Khandelwal, Zand, Matosich,
  Veeraraghavan, Michelena, Li, Jagadeesh, Huang, Chawla, Huang, Chen, Garg, A,
  Silva, Bell, Zhang, Guo, Yu, Moshkovich, Wehrstedt, Khabsa, Avalani, Bhatt,
  Mankus, Hasson, Lennie, Reso, Groshev, Naumov, Lathi, Keneally, Liu, Seltzer,
  Valko, Restrepo, Patel, Vyatskov, Samvelyan, Clark, Macey, Wang, Hermoso,
  Metanat, Rastegari, Bansal, Santhanam, Parks, White, Bawa, Singhal, Egebo,
  Usunier, Mehta, Laptev, Dong, Cheng, Chernoguz, Hart, Salpekar, Kalinli,
  Kent, Parekh, Saab, Balaji, Rittner, Bontrager, Roux, Dollar, Zvyagina,
  Ratanchandani, Yuvraj, Liang, Alao, Rodriguez, Ayub, Murthy, Nayani, Mitra,
  Parthasarathy, Li, Hogan, Battey, Wang, Howes, Rinott, Mehta, Siby, Bondu,
  Datta, Chugh, Hunt, Dhillon, Sidorov, Pan, Mahajan, Verma, Yamamoto,
  Ramaswamy, Lindsay, Lindsay, Feng, Lin, Zha, Patil, Shankar, Zhang, Zhang,
  Wang, Agarwal, Sajuyigbe, Chintala, Max, Chen, Kehoe, Satterfield,
  Govindaprasad, Gupta, Deng, Cho, Virk, Subramanian, Choudhury, Goldman,
  Remez, Glaser, Best, Koehler, Robinson, Li, Zhang, Matthews, Chou, Shaked,
  Vontimitta, Ajayi, Montanez, Mohan, Kumar, Mangla, Ionescu, Poenaru,
  Mihailescu, Ivanov, Li, Wang, Jiang, Bouaziz, Constable, Tang, Wu, Wang, Wu,
  Gao, Kleinman, Chen, Hu, Jia, Qi, Li, Zhang, Zhang, Adi, Nam, Wang, Zhao,
  Hao, Qian, Li, He, Rait, DeVito, Rosnbrick, Wen, Yang, Zhao, and
  Ma}]{grattafiori2024llama3}
Aaron Grattafiori, Abhimanyu Dubey, Abhinav Jauhri, Abhinav Pandey, Abhishek
  Kadian, Ahmad Al-Dahle, Aiesha Letman, Akhil Mathur, Alan Schelten, Alex
  Vaughan, Amy Yang, Angela Fan, Anirudh Goyal, Anthony Hartshorn, Aobo Yang,
  Archi Mitra, Archie Sravankumar, Artem Korenev, Arthur Hinsvark, and 540
  others. 2024.
\newblock \href {https://arxiv.org/abs/2407.21783} {The {Llama} 3 herd of
  models}.
\newblock \emph{arXiv preprint arXiv:2407.21783}.

\bibitem[{Guo et~al.(2025)Guo, Yang, Zhang, Song, Wang, Zhu, Xu, Zhang, Ma, Bi,
  Zhang, Yu, Wu, Wu, Gou, Shao, Li, Gao, Liu, Xue, Wang, Wu, Feng, Lu, Zhao,
  Deng, Ruan, Dai, Chen, Ji, Li, Lin, Dai, Luo, Hao, Chen, Li, Zhang, Xu, Ding,
  Gao, Qu, Li, Guo, Li, Chen, Yuan, Tu, Qiu, Li, Cai, Ni, Liang, Chen, Dong,
  Hu, You, Gao, Guan, Huang, Yu, Wang, Zhang, Zhao, Wang, Zhang, Xu, Xia,
  Zhang, Zhang, Tang, Zhou, Li, Wang, Li, Tian, Huang, Zhang, Wang, Chen, Du,
  Ge, Zhang, Pan, Wang, Chen, Jin, Chen, Lu, Zhou, Chen, Ye, Wang, Yu, Zhou,
  Pan, Li, Zhou, Wu, Yun, Pei, Sun, Wang, Zeng, Liu, Liang, Gao, Yu, Zhang,
  Xiao, An, Liu, Wang, Chen, Nie, Cheng, Liu, Xie, Liu, Yang, Li, Su, Lin, Li,
  Jin, Shen, Chen, Sun, Wang, Song, Zhou, Wang, Shan, Li, Wang, Wei, Zhang, Xu,
  Li, Zhao, Sun, Wang, Yu, Zhang, Shi, Xiong, He, Piao, Wang, Tan, Ma, Liu,
  Guo, Ou, Wang, Gong, Zou, He, Xiong, Luo, You, Liu, Zhou, Zhu, Huang, Li,
  Zheng, Zhu, Ma, Tang, Zha, Yan, Ren, Ren, Sha, Fu, Xu, Xie, Zhang, Hao, Ma,
  Yan, Wu, Gu, Zhu, Liu, Li, Xie, Song, Pan, Huang, Xu, Zhang, and
  Zhang}]{guo2025deepseekr1}
Daya Guo, Dejian Yang, Haowei Zhang, Junxiao Song, Peiyi Wang, Qihao Zhu,
  Runxin Xu, Ruoyu Zhang, Shirong Ma, Xiao Bi, Xiaokang Zhang, Xingkai Yu,
  Yu~Wu, Z.~F. Wu, Zhibin Gou, Zhihong Shao, Zhuoshu Li, Ziyi Gao, Aixin Liu,
  and 175 others. 2025.
\newblock \href {https://doi.org/10.1038/s41586-025-09422-z} {{DeepSeek-R1}
  incentivizes reasoning in {LLMs} through reinforcement learning}.
\newblock \emph{Nature}, 645(8081):633--638.

\bibitem[{Hendrycks et~al.(2021)Hendrycks, Burns, Kadavath, Arora, Basart,
  Tang, Song, and Steinhardt}]{hendrycks2021math}
Dan Hendrycks, Collin Burns, Saurav Kadavath, Akul Arora, Steven Basart, Eric
  Tang, Dawn Song, and Jacob Steinhardt. 2021.
\newblock \href
  {https://datasets-benchmarks-proceedings.neurips.cc/paper_files/paper/2021/file/be83ab3ecd0db773eb2dc1b0a17836a1-Paper-round2.pdf}
  {Measuring mathematical problem solving with the {MATH} dataset}.
\newblock In \emph{Proceedings of the Neural Information Processing Systems
  Track on Datasets and Benchmarks}, volume~1.

\bibitem[{Hinton et~al.(2015)Hinton, Vinyals, and Dean}]{hinton2015distilling}
Geoffrey Hinton, Oriol Vinyals, and Jeff Dean. 2015.
\newblock \href {https://arxiv.org/abs/1503.02531} {Distilling the knowledge in
  a neural network}.
\newblock \emph{arXiv preprint arXiv:1503.02531}.

\bibitem[{Ho et~al.(2023)Ho, Schmid, and Yun}]{ho2023reasoning}
Namgyu Ho, Laura Schmid, and Se-Young Yun. 2023.
\newblock \href {https://doi.org/10.18653/v1/2023.acl-long.830} {Large language
  models are reasoning teachers}.
\newblock In \emph{Proceedings of the 61st Annual Meeting of the Association
  for Computational Linguistics (Volume 1: Long Papers)}, pages 14852--14882,
  Toronto, Canada. Association for Computational Linguistics.

\bibitem[{Hu et~al.(2022)Hu, Shen, Wallis, Allen-Zhu, Li, Wang, Wang, and
  Chen}]{hu2021lora}
Edward~J. Hu, Yelong Shen, Phillip Wallis, Zeyuan Allen-Zhu, Yuanzhi Li, Shean
  Wang, Lu~Wang, and Weizhu Chen. 2022.
\newblock \href {https://openreview.net/forum?id=nZeVKeeFYf9} {{LoRA}: Low-rank
  adaptation of large language models}.
\newblock In \emph{International Conference on Learning Representations}.

\bibitem[{Hu et~al.(2026{\natexlab{a}})Hu, Ward, Icard, and
  Potts}]{hu2026transcoder}
Nathan Hu, Jake Ward, Thomas Icard, and Christopher Potts. 2026{\natexlab{a}}.
\newblock \href {https://arxiv.org/abs/2602.20904} {Transcoder adapters for
  reasoning-model diffing}.
\newblock \emph{arXiv preprint arXiv:2602.20904}.

\bibitem[{Hu et~al.(2026{\natexlab{b}})Hu, Xiao, Song, Jiang, Li, Xiong, Liu,
  Lin, Su, Hu, Ding, Xiao, and Poovendran}]{hu2026strongest}
Zhengyu Hu, Zheyuan Xiao, Linxin Song, Fengqing Jiang, Yuetai Li, Zhihan Xiong,
  Yue Liu, Junhao Lin, Yao Su, Lijie Hu, Kaize Ding, Teng Xiao, and Radha
  Poovendran. 2026{\natexlab{b}}.
\newblock \href {https://arxiv.org/abs/2605.26872} {When the strongest teacher
  is not the best teacher: Student-centric answer selection}.
\newblock \emph{arXiv preprint arXiv:2605.26872}.

\bibitem[{Ilharco et~al.(2023)Ilharco, Ribeiro, Wortsman, Gururangan, Schmidt,
  Hajishirzi, and Farhadi}]{ilharco2023task}
Gabriel Ilharco, Marco~Tulio Ribeiro, Mitchell Wortsman, Suchin Gururangan,
  Ludwig Schmidt, Hannaneh Hajishirzi, and Ali Farhadi. 2023.
\newblock \href {https://openreview.net/forum?id=6t0Kwf8-jrj} {Editing models
  with task arithmetic}.
\newblock In \emph{The Eleventh International Conference on Learning
  Representations}.

\bibitem[{Jayabahu and Adeleke(2026)}]{jayabahu2026halt}
Dylan Jayabahu and Tinuade Adeleke. 2026.
\newblock \href {https://arxiv.org/abs/2608.28859} {The halt vector:
  Internalizing a causal steering intervention for efficient reasoning}.
\newblock \emph{arXiv preprint arXiv:2608.28859}.

\bibitem[{Jiang et~al.(2025)Jiang, Lu, Lin, Han, and Sun}]{jiang2025teach}
Wangyi Jiang, Yaojie Lu, Hongyu Lin, Xianpei Han, and Le~Sun. 2025.
\newblock \href {https://doi.org/10.18653/v1/2025.emnlp-main.376} {Teach small
  models to reason by curriculum distillation}.
\newblock In \emph{Proceedings of the 2025 Conference on Empirical Methods in
  Natural Language Processing}, pages 7412--7422, Suzhou, China. Association
  for Computational Linguistics.

\bibitem[{Jiralerspong and Bricken(2026)}]{jiralerspong2026crossarch}
Thomas Jiralerspong and Trenton Bricken. 2026.
\newblock \href {https://arxiv.org/abs/2602.11729} {Cross-architecture model
  diffing with crosscoders: Unsupervised discovery of differences between
  {LLMs}}.
\newblock \emph{arXiv preprint arXiv:2602.11729}.

\bibitem[{Kim and Rush(2016)}]{kim2016sequence}
Yoon Kim and Alexander~M. Rush. 2016.
\newblock \href {https://doi.org/10.18653/v1/D16-1139} {Sequence-level
  knowledge distillation}.
\newblock In \emph{Proceedings of the 2016 Conference on Empirical Methods in
  Natural Language Processing}, pages 1317--1327, Austin, Texas. Association
  for Computational Linguistics.

\bibitem[{{Kimi Team} et~al.(2025){Kimi Team}, Du, Gao, Xing, Jiang, Chen, Li,
  Xiao, Du, Liao, Tang, Wang, Zhang, Yuan, Lu, Tang, Sung, Wei, Lai, Guo, Zhu,
  Ding, Hu, Yang, Zhang, Yao, Zhao, Lu, Li, Yu, Gao, Zheng, Yuan, Chen, Guo,
  Su, Wang, Zhao, Zhang, Liu, Yan, Wu, Shi, Ye, Yu, Dong, Zhang, Ma, Pan, Gong,
  Liu, Ma, Wei, Cao, Huang, Jiang, Gao, Xiong, He, Huang, Xu, Wu, He, Wei, Jia,
  Wu, Xu, Zu, Zhou, Pan, Charles, Li, Hu, Liu, Chen, Wang, Liu, Qin, Liu, Yang,
  Bao, Du, Wu, Wang, Zhou, Wang, Li, Zhu, Zhang, Wang, Yang, Huang, Huang, Xu,
  Yang, and Lin}]{kimi2025k15}
{Kimi Team}, Angang Du, Bofei Gao, Bowei Xing, Changjiu Jiang, Cheng Chen,
  Cheng Li, Chenjun Xiao, Chenzhuang Du, Chonghua Liao, Chuning Tang, Congcong
  Wang, Dehao Zhang, Enming Yuan, Enzhe Lu, Fengxiang Tang, Flood Sung, Guangda
  Wei, Guokun Lai, and 77 others. 2025.
\newblock \href {https://arxiv.org/abs/2501.12599} {{Kimi} k1.5: Scaling
  reinforcement learning with {LLMs}}.
\newblock \emph{arXiv preprint arXiv:2501.12599}.

\bibitem[{Klagges et~al.(2025)Klagges, Dahlke, Klemm, Merkel, Klingmann, Reiss,
  and Zecha}]{klagges2025assembly}
Henrik Klagges, Robert Dahlke, Fabian Klemm, Benjamin Merkel, Daniel Klingmann,
  David~A. Reiss, and Dan Zecha. 2025.
\newblock \href {https://arxiv.org/abs/2506.14794} {Assembly of experts:
  Linear-time construction of the {Chimera} {LLM} variants with emergent and
  adaptable behaviors}.
\newblock \emph{arXiv preprint arXiv:2506.14794}.

\bibitem[{Kwon et~al.(2023)Kwon, Li, Zhuang, Sheng, Zheng, Yu, Gonzalez, Zhang,
  and Stoica}]{kwon2023vllm}
Woosuk Kwon, Zhuohan Li, Siyuan Zhuang, Ying Sheng, Lianmin Zheng, Cody~Hao Yu,
  Joseph~E. Gonzalez, Hao Zhang, and Ion Stoica. 2023.
\newblock \href {https://doi.org/10.1145/3600006.3613165} {Efficient memory
  management for large language model serving with {PagedAttention}}.
\newblock In \emph{Proceedings of the 29th Symposium on Operating Systems
  Principles}, pages 611--626. ACM.

\bibitem[{Lan et~al.(2025)Lan, Zheng, Cao, and Li}]{lan2025spectrum}
Xiaochong Lan, Yu~Zheng, Shiteng Cao, and Yong Li. 2025.
\newblock \href {https://arxiv.org/abs/2509.22034} {The thinking spectrum: An
  empirical study of tunable reasoning in {LLMs} through model merging}.
\newblock \emph{arXiv preprint arXiv:2509.22034}.

\bibitem[{Lan et~al.(2026)Lan, Hu, Zheng, Zhang, Wang, Dai, and
  Li}]{lan2026midistillation}
Yangsong Lan, Renkai Hu, HongKai Zheng, Bo~Zhang, Renzhi Wang, Hongliang Dai,
  and Piji Li. 2026.
\newblock \href {https://arxiv.org/abs/2608.29623} {{MI-Distillation}:
  Selecting from model-interpolated instruct-reasoning data spectrum for
  chain-of-thought distillation}.
\newblock \emph{arXiv preprint arXiv:2608.29623}.

\bibitem[{Lee et~al.(2025)Lee, Kwon, and won Hwang}]{lee2025goldswitch}
Jaeseong Lee, Dayoung Kwon, and Seung won Hwang. 2025.
\newblock \href {https://arxiv.org/abs/2510.06750} {{Gold-Switch}:
  Training-free superposition of slow- and fast- thinking {LLMs}}.
\newblock \emph{arXiv preprint arXiv:2510.06750}.

\bibitem[{Li et~al.(2025)Li, Yue, Xu, Jiang, Niu, Lin, Ramasubramanian, and
  Poovendran}]{li2025small}
Yuetai Li, Xiang Yue, Zhangchen Xu, Fengqing Jiang, Luyao Niu, Bill~Yuchen Lin,
  Bhaskar Ramasubramanian, and Radha Poovendran. 2025.
\newblock \href {https://doi.org/10.18653/v1/2025.findings-acl.1301} {Small
  models struggle to learn from strong reasoners}.
\newblock In \emph{Findings of the Association for Computational Linguistics:
  ACL 2025}, pages 25366--25394, Vienna, Austria. Association for Computational
  Linguistics.

\bibitem[{Lightman et~al.(2024)Lightman, Kosaraju, Burda, Edwards, Baker, Lee,
  Leike, Schulman, Sutskever, and Cobbe}]{lightman2023verify}
Hunter Lightman, Vineet Kosaraju, Yura Burda, Harri Edwards, Bowen Baker, Teddy
  Lee, Jan Leike, John Schulman, Ilya Sutskever, and Karl Cobbe. 2024.
\newblock \href {https://openreview.net/forum?id=v8L0pN6EOi} {Let's verify step
  by step}.
\newblock In \emph{The Twelfth International Conference on Learning
  Representations}.

\bibitem[{Lion et~al.(2025)Lion, Zhang, Li, and He}]{lion2025polar}
Kai Lion, Liang Zhang, Bingcong Li, and Niao He. 2025.
\newblock \href {https://doi.org/10.52202/085713-4996} {{PoLAR}:
  Polar-decomposed low-rank adapter representation}.
\newblock In \emph{Advances in Neural Information Processing Systems},
  volume~38, pages 149313--149358. Curran Associates, Inc.

\bibitem[{Luo et~al.(2025)Luo, Li, Huang, and Lu}]{luo2025valley}
Renjie Luo, Jiaxi Li, Chen Huang, and Wei Lu. 2025.
\newblock \href {https://doi.org/10.18653/v1/2025.emnlp-main.251} {Through the
  valley: Path to effective long {C}o{T} training for small language models}.
\newblock In \emph{Proceedings of the 2025 Conference on Empirical Methods in
  Natural Language Processing}, pages 4972--4992, Suzhou, China. Association
  for Computational Linguistics.

\bibitem[{Ma et~al.(2026{\natexlab{a}})Ma, Liang, Chen, and
  Sojoudi}]{ma2026sae}
George Ma, Zhongyuan Liang, Irene~Y. Chen, and Somayeh Sojoudi.
  2026{\natexlab{a}}.
\newblock \href {https://proceedings.mlr.press/v306/ma26v.html} {Do sparse
  autoencoders identify reasoning features in language models?}
\newblock In \emph{Proceedings of the 43rd International Conference on Machine
  Learning}, volume 306 of \emph{Proceedings of Machine Learning Research},
  pages 84641--84688. PMLR.

\bibitem[{Ma et~al.(2026{\natexlab{b}})Ma, Cao, Fan, Yang, and
  Liu}]{ma2026spectral}
Hongbo Ma, Sansheng Cao, Jiajun Fan, Bangji Yang, and Ge~Liu.
  2026{\natexlab{b}}.
\newblock \href {https://arxiv.org/abs/2609.37976} {{$S^3$}: Spectral
  null-space swap makes reasoning models efficient}.
\newblock \emph{arXiv preprint arXiv:2609.37976}.

\bibitem[{Ma et~al.(2025{\natexlab{a}})Ma, He, Snell, Griggs, Min, and
  Zaharia}]{ma2025nothinking}
Wenjie Ma, Jingxuan He, Charlie Snell, Tyler Griggs, Sewon Min, and Matei
  Zaharia. 2025{\natexlab{a}}.
\newblock \href {https://arxiv.org/abs/2504.09858} {Reasoning models can be
  effective without thinking}.
\newblock \emph{arXiv preprint arXiv:2504.09858}.

\bibitem[{Ma et~al.(2025{\natexlab{b}})Ma, Wan, Yu, Fang, and
  Wang}]{ma2025cotvalve}
Xinyin Ma, Guangnian Wan, Runpeng Yu, Gongfan Fang, and Xinchao Wang.
  2025{\natexlab{b}}.
\newblock \href {https://doi.org/10.18653/v1/2025.acl-long.300}
  {{C}o{T}-{V}alve: Length-compressible chain-of-thought tuning}.
\newblock In \emph{Proceedings of the 63rd Annual Meeting of the Association
  for Computational Linguistics (Volume 1: Long Papers)}, pages 6025--6035,
  Vienna, Austria. Association for Computational Linguistics.

\bibitem[{Magister et~al.(2023)Magister, Mallinson, Adamek, Malmi, and
  Severyn}]{magister2023teaching}
Lucie~Charlotte Magister, Jonathan Mallinson, Jakub Adamek, Eric Malmi, and
  Aliaksei Severyn. 2023.
\newblock \href {https://doi.org/10.18653/v1/2023.acl-short.151} {Teaching
  small language models to reason}.
\newblock In \emph{Proceedings of the 61st Annual Meeting of the Association
  for Computational Linguistics (Volume 2: Short Papers)}, pages 1773--1781,
  Toronto, Canada. Association for Computational Linguistics.

\bibitem[{{Meta AI}(2024)}]{meta2024llama32}
{Meta AI}. 2024.
\newblock \href
  {https://ai.meta.com/blog/llama-3-2-connect-2024-vision-edge-mobile-devices/}
  {{Llama} 3.2: Revolutionizing edge {AI} and vision with open, customizable
  models}.

\bibitem[{Muennighoff et~al.(2025)Muennighoff, Yang, Shi, Li, Fei-Fei,
  Hajishirzi, Zettlemoyer, Liang, Cand{\`e}s, and
  Hashimoto}]{muennighoff2025s1}
Niklas Muennighoff, Zitong Yang, Weijia Shi, Xiang~Lisa Li, Li~Fei-Fei,
  Hannaneh Hajishirzi, Luke Zettlemoyer, Percy Liang, Emmanuel Cand{\`e}s, and
  Tatsunori Hashimoto. 2025.
\newblock \href {https://doi.org/10.18653/v1/2025.emnlp-main.1025} {s1: Simple
  test-time scaling}.
\newblock In \emph{Proceedings of the 2025 Conference on Empirical Methods in
  Natural Language Processing}, pages 20275--20321, Suzhou, China. Association
  for Computational Linguistics.

\bibitem[{Panigrahi et~al.(2023)Panigrahi, Saunshi, Zhao, and
  Arora}]{panigrahi2023skill}
Abhishek Panigrahi, Nikunj Saunshi, Haoyu Zhao, and Sanjeev Arora. 2023.
\newblock \href {https://proceedings.mlr.press/v202/panigrahi23a.html}
  {Task-specific skill localization in fine-tuned language models}.
\newblock In \emph{Proceedings of the 40th International Conference on Machine
  Learning}, volume 202 of \emph{Proceedings of Machine Learning Research},
  pages 27011--27033. PMLR.

\bibitem[{Paszke et~al.(2019)Paszke, Gross, Massa, Lerer, Bradbury, Chanan,
  Killeen, Lin, Gimelshein, Antiga, Desmaison, K{\"o}pf, Yang, DeVito, Raison,
  Tejani, Chilamkurthy, Steiner, Fang, Bai, and Chintala}]{paszke2019pytorch}
Adam Paszke, Sam Gross, Francisco Massa, Adam Lerer, James Bradbury, Gregory
  Chanan, Trevor Killeen, Zeming Lin, Natalia Gimelshein, Luca Antiga, Alban
  Desmaison, Andreas K{\"o}pf, Edward Yang, Zach DeVito, Martin Raison, Alykhan
  Tejani, Sasank Chilamkurthy, Benoit Steiner, Lu~Fang, and 2 others. 2019.
\newblock \href
  {https://proceedings.neurips.cc/paper_files/paper/2019/file/bdbca288fee7f92f2bfa9f7012727740-Paper.pdf}
  {{PyTorch}: An imperative style, high-performance deep learning library}.
\newblock In \emph{Advances in Neural Information Processing Systems},
  volume~32. Curran Associates, Inc.

\bibitem[{{Qwen} et~al.(2025){Qwen}, Yang, Yang, Zhang, Hui, Zheng, Yu, Li,
  Liu, Huang, Wei, Lin, Yang, Tu, Zhang, Yang, Yang, Zhou, Lin, Dang, Lu, Bao,
  Yang, Yu, Li, Xue, Zhang, Zhu, Men, Lin, Li, Tang, Xia, Ren, Ren, Fan, Su,
  Zhang, Wan, Liu, Cui, Zhang, and Qiu}]{qwen2025qwen25}
{Qwen}, An~Yang, Baosong Yang, Beichen Zhang, Binyuan Hui, Bo~Zheng, Bowen Yu,
  Chengyuan Li, Dayiheng Liu, Fei Huang, Haoran Wei, Huan Lin, Jian Yang,
  Jianhong Tu, Jianwei Zhang, Jianxin Yang, Jiaxi Yang, Jingren Zhou, Junyang
  Lin, and 24 others. 2025.
\newblock \href {https://arxiv.org/abs/2412.15115} {{Qwen2.5} technical
  report}.
\newblock \emph{arXiv preprint arXiv:2412.15115}.

\bibitem[{Rein et~al.(2024)Rein, Hou, Stickland, Petty, Pang, Dirani, Michael,
  and Bowman}]{rein2023gpqa}
David Rein, Betty~Li Hou, Asa~Cooper Stickland, Jackson Petty, Richard~Yuanzhe
  Pang, Julien Dirani, Julian Michael, and Samuel~R. Bowman. 2024.
\newblock \href {https://openreview.net/forum?id=Ti67584b98} {{GPQA}: A
  graduate-level {Google}-proof {Q\&A} benchmark}.
\newblock In \emph{First Conference on Language Modeling}.

\bibitem[{Saini and Pandey(2026)}]{saini2026thinkswitch}
Dhruv Saini and Rohan Pandey. 2026.
\newblock \href {https://arxiv.org/abs/2606.01080} {{ThinkSwitch}: Context
  distillation with {LoRA} and weight interpolation for specific-purpose
  reasoning tasks}.
\newblock \emph{arXiv preprint arXiv:2606.01080}.

\bibitem[{Sheng et~al.(2025)Sheng, Zhang, Wu, Zhao, Shen, Zhang, Wang, and
  Chua}]{sheng2025planning}
Leheng Sheng, An~Zhang, Zijian Wu, Weixiang Zhao, Changshuo Shen, Yi~Zhang,
  Xiang Wang, and Tat-Seng Chua. 2025.
\newblock \href {https://doi.org/10.52202/085713-2115} {On reasoning strength
  planning in large reasoning models}.
\newblock In \emph{Advances in Neural Information Processing Systems},
  volume~38, pages 62978--63012. Curran Associates, Inc.

\bibitem[{Shuttleworth et~al.(2025)Shuttleworth, Andreas, Torralba, and
  Sharma}]{shuttleworth2025lora}
Reece Shuttleworth, Jacob Andreas, Antonio Torralba, and Pratyusha Sharma.
  2025.
\newblock \href {https://doi.org/10.52202/085713-5808} {{LoRA} vs full
  fine-tuning: An illusion of equivalence}.
\newblock In \emph{Advances in Neural Information Processing Systems},
  volume~38, pages 174627--174662. Curran Associates, Inc.

\bibitem[{Sun et~al.(2025)Sun, Yan, and Weng}]{sun2025thinkedit}
Chung-En Sun, Ge~Yan, and Tsui-Wei Weng. 2025.
\newblock \href {https://doi.org/10.18653/v1/2025.emnlp-main.861}
  {{T}hink{E}dit: Interpretable weight editing to mitigate overly short
  thinking in reasoning models}.
\newblock In \emph{Proceedings of the 2025 Conference on Empirical Methods in
  Natural Language Processing}, pages 17001--17025, Suzhou, China. Association
  for Computational Linguistics.

\bibitem[{{Team OLMo} et~al.(2025){Team OLMo}, Walsh, Soldaini, Groeneveld, Lo,
  Arora, Bhagia, Gu, Huang, Jordan, Lambert, Schwenk, Tafjord, Anderson,
  Atkinson, Brahman, Clark, Dasigi, Dziri, Ettinger, Guerquin, Heineman,
  Ivison, Koh, Liu, Malik, Merrill, Miranda, Morrison, Murray, Nam, Poznanski,
  Pyatkin, Rangapur, Schmitz, Skjonsberg, Wadden, Wilhelm, Wilson, Zettlemoyer,
  Farhadi, Smith, and Hajishirzi}]{olmo2025olmo2}
{Team OLMo}, Pete Walsh, Luca Soldaini, Dirk Groeneveld, Kyle Lo, Shane Arora,
  Akshita Bhagia, Yuling Gu, Shengyi Huang, Matt Jordan, Nathan Lambert, Dustin
  Schwenk, Oyvind Tafjord, Taira Anderson, David Atkinson, Faeze Brahman,
  Christopher Clark, Pradeep Dasigi, Nouha Dziri, and 24 others. 2025.
\newblock \href {https://arxiv.org/abs/2501.00656} {2 {OLMo} 2 furious}.
\newblock \emph{arXiv preprint arXiv:2501.00656}.

\bibitem[{Venhoff et~al.(2025)Venhoff, Arcuschin, Torr, Conmy, and
  Nanda}]{venhoff2025steering}
Constantin Venhoff, Iv{\'a}n Arcuschin, Philip Torr, Arthur Conmy, and Neel
  Nanda. 2025.
\newblock \href {https://arxiv.org/abs/2506.18167} {Understanding reasoning in
  thinking language models via steering vectors}.
\newblock \emph{arXiv preprint arXiv:2506.18167}.

\bibitem[{Venhoff et~al.(2026)Venhoff, Arcuschin, Torr, Conmy, and
  Nanda}]{venhoff2026base}
Constantin Venhoff, Iv{\'a}n Arcuschin, Philip Torr, Arthur Conmy, and Neel
  Nanda. 2026.
\newblock \href {https://proceedings.mlr.press/v306/venhoff26a.html} {Base
  models know how to reason, thinking models learn when}.
\newblock In \emph{Proceedings of the 43rd International Conference on Machine
  Learning}, volume 306 of \emph{Proceedings of Machine Learning Research},
  pages 123688--123732. PMLR.

\bibitem[{Ward et~al.(2025{\natexlab{a}})Ward, Lin, Venhoff, and
  Nanda}]{ward2025repurposes}
Jake Ward, Chuqiao Lin, Constantin Venhoff, and Neel Nanda. 2025{\natexlab{a}}.
\newblock \href {https://arxiv.org/abs/2507.12638} {Reasoning-finetuning
  repurposes latent representations in base models}.
\newblock \emph{arXiv preprint arXiv:2507.12638}.

\bibitem[{Ward et~al.(2025{\natexlab{b}})Ward, Riechers, and
  Shai}]{ward2025rank1}
Jake Ward, Paul Riechers, and Adam Shai. 2025{\natexlab{b}}.
\newblock \href {https://arxiv.org/abs/2511.06739} {Rank-1 {LoRAs} encode
  interpretable reasoning signals}.
\newblock \emph{arXiv preprint arXiv:2511.06739}.

\bibitem[{Wei et~al.(2024)Wei, Huang, Huang, Xie, Qi, Xia, Mittal, Wang, and
  Henderson}]{wei2024brittleness}
Boyi Wei, Kaixuan Huang, Yangsibo Huang, Tinghao Xie, Xiangyu Qi, Mengzhou Xia,
  Prateek Mittal, Mengdi Wang, and Peter Henderson. 2024.
\newblock \href {https://proceedings.mlr.press/v235/wei24f.html} {Assessing the
  brittleness of safety alignment via pruning and low-rank modifications}.
\newblock In \emph{Proceedings of the 41st International Conference on Machine
  Learning}, volume 235 of \emph{Proceedings of Machine Learning Research},
  pages 52588--52610. PMLR.

\bibitem[{Wolf et~al.(2020)Wolf, Debut, Sanh, Chaumond, Delangue, Moi, Cistac,
  Rault, Louf, Funtowicz, Davison, Shleifer, von Platen, Ma, Jernite, Plu, Xu,
  Le~Scao, Gugger, Drame, Lhoest, and Rush}]{wolf2020transformers}
Thomas Wolf, Lysandre Debut, Victor Sanh, Julien Chaumond, Clement Delangue,
  Anthony Moi, Pierric Cistac, Tim Rault, R{\'e}mi Louf, Morgan Funtowicz, Joe
  Davison, Sam Shleifer, Patrick von Platen, Clara Ma, Yacine Jernite, Julien
  Plu, Canwen Xu, Teven Le~Scao, Sylvain Gugger, and 3 others. 2020.
\newblock \href {https://doi.org/10.18653/v1/2020.emnlp-demos.6} {Transformers:
  State-of-the-art natural language processing}.
\newblock In \emph{Proceedings of the 2020 Conference on Empirical Methods in
  Natural Language Processing: System Demonstrations}, pages 38--45, Online.
  Association for Computational Linguistics.

\bibitem[{Wortsman et~al.(2022)Wortsman, Ilharco, Gadre, Roelofs,
  Gontijo-Lopes, Morcos, Namkoong, Farhadi, Carmon, Kornblith, and
  Schmidt}]{wortsman2022soups}
Mitchell Wortsman, Gabriel Ilharco, Samir~Yitzhak Gadre, Rebecca Roelofs,
  Raphael Gontijo-Lopes, Ari~S. Morcos, Hongseok Namkoong, Ali Farhadi, Yair
  Carmon, Simon Kornblith, and Ludwig Schmidt. 2022.
\newblock \href {https://proceedings.mlr.press/v162/wortsman22a.html} {Model
  soups: averaging weights of multiple fine-tuned models improves accuracy
  without increasing inference time}.
\newblock In \emph{Proceedings of the 39th International Conference on Machine
  Learning}, volume 162 of \emph{Proceedings of Machine Learning Research},
  pages 23965--23998. PMLR.

\bibitem[{Wu et~al.(2025)Wu, Yao, Liu, Liu, Fu, Han, Li, Zhen, Zhong, and
  Yuan}]{wu2025merging}
Han Wu, Yuxuan Yao, Shuqi Liu, Zehua Liu, Xiaojin Fu, Xiongwei Han, Xing Li,
  Hui-Ling Zhen, Tao Zhong, and Mingxuan Yuan. 2025.
\newblock \href {https://arxiv.org/abs/2503.20641} {Unlocking efficient
  long-to-short {LLM} reasoning with model merging}.
\newblock \emph{arXiv preprint arXiv:2503.20641}.

\bibitem[{Wu et~al.(2026{\natexlab{a}})Wu, Zhang, Dong, Xi, Zhao, Jin, Fan,
  Zhou, Lv, Zhang, Fu, Liu, Zhang, and Zhang}]{wu2026memorization}
Mingqi Wu, Zhihao Zhang, Qiaole Dong, Zhiheng Xi, Jun Zhao, Senjie Jin, Xiaoran
  Fan, Yuhao Zhou, Huijie Lv, Ming Zhang, Yanwei Fu, Qin Liu, Songyang Zhang,
  and Qi~Zhang. 2026{\natexlab{a}}.
\newblock \href {https://doi.org/10.1609/aaai.v40i40.40687} {Reasoning or
  memorization? {U}nreliable results of reinforcement learning due to data
  contamination}.
\newblock \emph{Proceedings of the AAAI Conference on Artificial Intelligence},
  40(40):33944--33952.

\bibitem[{Wu et~al.(2026{\natexlab{b}})Wu, Yang, Liu, Wang, and
  Wong}]{wu2026interpolation}
Taiqiang Wu, Runming Yang, Tao Liu, Jiahao Wang, and Ngai Wong.
  2026{\natexlab{b}}.
\newblock \href {https://doi.org/10.18653/v1/2026.acl-long.389} {Revisiting
  model interpolation for efficient reasoning}.
\newblock In \emph{Proceedings of the 64th Annual Meeting of the {A}ssociation
  for {C}omputational {L}inguistics (Volume 1: Long Papers)}, pages 8624--8638,
  San Diego, California, United States. Association for Computational
  Linguistics.

\bibitem[{Yang et~al.(2025)Yang, Li, Yang, Zhang, Hui, Zheng, Yu, Gao, Huang,
  Lv, Zheng, Liu, Zhou, Huang, Hu, Ge, Wei, Lin, Tang, Yang, Tu, Zhang, Yang,
  Yang, Zhou, Zhou, Lin, Dang, Bao, Yang, Yu, Deng, Li, Xue, Li, Zhang, Wang,
  Zhu, Men, Gao, Liu, Luo, Li, Tang, Yin, Ren, Wang, Zhang, Ren, Fan, Su,
  Zhang, Zhang, Wan, Liu, Wang, Cui, Zhang, Zhou, and Qiu}]{yang2025qwen3}
An~Yang, Anfeng Li, Baosong Yang, Beichen Zhang, Binyuan Hui, Bo~Zheng, Bowen
  Yu, Chang Gao, Chengen Huang, Chenxu Lv, Chujie Zheng, Dayiheng Liu, Fan
  Zhou, Fei Huang, Feng Hu, Hao Ge, Haoran Wei, Huan Lin, Jialong Tang, and 41
  others. 2025.
\newblock \href {https://arxiv.org/abs/2505.09388} {{Qwen3} technical report}.
\newblock \emph{arXiv preprint arXiv:2505.09388}.

\bibitem[{Yang et~al.(2024)Yang, Zhang, Hui, Gao, Yu, Li, Liu, Tu, Zhou, Lin,
  Lu, Xue, Lin, Liu, Ren, and Zhang}]{yang2024qwen25math}
An~Yang, Beichen Zhang, Binyuan Hui, Bofei Gao, Bowen Yu, Chengpeng Li,
  Dayiheng Liu, Jianhong Tu, Jingren Zhou, Junyang Lin, Keming Lu, Mingfeng
  Xue, Runji Lin, Tianyu Liu, Xingzhang Ren, and Zhenru Zhang. 2024.
\newblock \href {https://arxiv.org/abs/2409.12122} {{Qwen2.5-Math} technical
  report: Toward mathematical expert model via self-improvement}.
\newblock \emph{arXiv preprint arXiv:2409.12122}.

\bibitem[{Yang et~al.(2026)Yang, Si, Tian, Liu, Yao, Qin, Lin, Wang, and
  Wang}]{yang2026system}
Chenxu Yang, Qingyi Si, Chong Tian, Xiyu Liu, Dingyu Yao, Chuanyu Qin, Zheng
  Lin, Weiping Wang, and Jiaqi Wang. 2026.
\newblock \href {https://doi.org/10.24963/ijcai.2026/668} {System 1\&2 synergy
  via dynamic model interpolation}.
\newblock In \emph{Proceedings of the Thirty-Fifth International Joint
  Conference on Artificial Intelligence, {IJCAI-26}}, pages 6000--6008.
  International Joint Conferences on Artificial Intelligence Organization.

\bibitem[{Yeo et~al.(2025)Yeo, Tong, Niu, Neubig, and
  Yue}]{yeo2025demystifying}
Edward Yeo, Yuxuan Tong, Morry Niu, Graham Neubig, and Xiang Yue. 2025.
\newblock \href {https://proceedings.mlr.press/v267/yang25ae.html}
  {Demystifying long chain-of-thought reasoning}.
\newblock In \emph{Proceedings of the 42nd International Conference on Machine
  Learning}, volume 267 of \emph{Proceedings of Machine Learning Research},
  pages 71177--71209. PMLR.

\end{thebibliography}

\clearpage
\appendix
\section{Experimental Details}
\label{app:details}

\paragraph{Data and training.}
From 4,000 MATH training problems without MATH-500 overlap, we keep those solved completely and correctly within 8,192 tokens at all six $\lambda$ (for pair A also at $\lambda=0.1$ and 0.3), holding out 300 for validation. OLMo-2 has a 4,096-token context, so 9\%, 29\% and 6\% of its $\lambda=0$ training solutions (pairs A, B, C) are truncated. Adapters are trained on seven linear-layer types with batch size 32 and learning rate $2\times10^{-4}$ for 3 epochs. We use PyTorch \citep{paszke2019pytorch}, Transformers \citep{wolf2020transformers} and PEFT,\footnote{\url{https://github.com/huggingface/peft}} and sample with vLLM \citep{kwon2023vllm} at $T=0.6$ and top-p 0.95 up to 8,192 tokens for teachers and 7,680 for students (3,840 for OLMo). GSM8K uses its first 200 problems and GPQA-Diamond all 198, and untrained students vary three generation seeds. Code, results, adapters and outputs are public (\S\ref{sec:setup}).

\paragraph{Measuring thinking.}
Markers are \emph{wait}, \emph{hmm}, \emph{actually}, \emph{but let me}, \emph{let me check}, \emph{recheck}, \emph{verify}, \emph{reconsider}, \emph{alternatively}, \emph{on second thought}, \emph{let me re}, \emph{double-check}, \emph{double check}, \emph{is that right} and \emph{hold on}. We count them per 1,000 words in the first of two responses per problem, and early markers in its first 300 words. Think tags and template tokens are stripped, so the students never see \texttt{</think>}. A public SAE for the base Llama-3.2-1B\footnote{\url{https://huggingface.co/EleutherAI/sae-Llama-3.2-1B-131k}} gives 83 features of the layer-11 MLP output, which the adapters also edit, that fire more on $\lambda=0$ than on $\lambda=1$ teacher solutions. They survive inserting markers \citep{ma2026sae} and deleting them. The SAE leaves 0.56--0.59 of the variance unexplained and raises the next-token loss by 0.01--0.04 when spliced in, so it is only an auxiliary read-out.

\paragraph{Tests and regions.}
The permutation test averages each pair's three student correlations and permutes each pair's teacher order ($120^3$ orderings). It treats a pair's five intervals as exchangeable under the null. With two adapters, Eq.~\eqref{eq:4} gives a paired direction the eigenvalue $(1+\cos\theta)/2$, so the 0.5 cut keeps nearly all. The two-seed region therefore keeps only directions within $60^\circ$ ($\mu > 0.75$). Rebuilt at cuts 0.4 and 0.6, the region has 9.5--11.9 and 2.8--4.9 directions. Pre-switch coverage stays 2.9--9.7 times post-switch coverage, and every build keeps at least one direction per module. Region sizes come from all pre-switch adapters weighted by $\lVert \Delta W \rVert_F$ (5.4--7.7) or unweighted (6.4--8.0), and for the edits from those without the edited adapter, unweighted (7.2--8.4).

\paragraph{Edits.}
Removal sets $B \leftarrow B - V(V^\top B)$. Its controls take energy from the strongest directions orthogonal to $V$ or from the whole update, matched in $\lVert B \rVert^2$ per module, or from the largest singular components, taking at least its $\lVert \Delta W \rVert^2$. In some modules the region holds more of $\lVert B \rVert^2$ than the rest, and there the control removes every outside direction (Table~\ref{tab:energy}). Energy-matched top removal takes 1.6--2.5 directions per module, a quarter to a third of the region's, mostly inside it (57--74\%, against at most 0.3\% by chance), and they span 14--23\% of it. The post-switch region keeps the region's number of top eigenvectors of the mean projector from the switch $\lambda$ on, including pair C's $\lambda=0.6$, where the teacher still thinks in 27\% of responses. Gold-Switch-style removal takes the largest rank $r$ whose tail energy is at least $\tau=0.6$. \citet{lee2025goldswitch} state the minimum such rank, which is always 1, so the largest is our reading, applied to a student adapter rather than an LRM's difference from its base. Grafts pair same-seed adapters, and the think-skip prompt only says that thinking is finished.

\section{Measuring the Switch}
\label{app:measure}

\paragraph{Normalizing $R$.}
In A/Qwen the data-half overlap $B(\lambda)$ dividing $R$ falls from 0.38 at $\lambda=0.2$ to 0.07 at 0.4 and 0.04 at 1. $R$ therefore stays 0.62--0.74 after the switch, while elsewhere $B(\lambda)$ ends at 0.14--0.48. Dividing instead by $\sqrt{B(0)B(\lambda)}$ puts the largest drop at the switch in all nine cells, and pair A's first interval drops 0.32--0.33 against 0.45--0.52 at the switch. The drop also remains across seeds in every pair, and Figure~\ref{fig:switch} (bottom) gives the share of the $\lambda=0$ region kept at every $\lambda$.

\paragraph{Interval classes.}
Table~\ref{tab:length} classifies intervals by the teacher alone. An interval holds thinking when the think rate changes by under 15 points and stays at 50\% or more, and the teacher's end markers stay at least twice their $\lambda=1$ level. It is a switch when the rate falls by at least 50 points, or by at least 15 from 15\% or more right after a switch. Held intervals move the rate by at most 10.0 points, and first switches drop it by 62.0--90.5. Every threshold between 26.8 and 62.0 gives the same first steps. The half-data reference is approximate, as half-data adapters may cover less. Per-unit losses assume no loss without a length change (held B and C intervals keep at least 0.78).

\paragraph{Loss per unit of length.}
An interval's loss is $-\ln$ of its kept share (with $1-$kept, pair A's switch/held ratio is 1.47--1.63, against 1.19--1.44 for the whole update). Per unit of log-length drop, pair A's switch loses 2.46--2.62 times as much as its held interval (Table~\ref{tab:lenbody}). In pairs B and C the mean loss is also lower over held intervals than at switch steps. Pair B's 0.6 to 0.8 interval loses more than its own switch (Table~\ref{tab:lenbody}), and there the $\lambda=0$ region loses close to its switch rate. The teacher still writes 5.0 to 4.2 markers per 1,000 words there (3.7 at $\lambda=1$), so B's short-answer end is partly reflective. Pair A's switch is also its largest drop in log teacher markers (1.59 against 1.14), and per unit of that drop it loses 1.05--1.33 against 0.47--0.61. Scoring each end seed separately gives 2.43--2.70. A region built from the other two seeds at cut 0.75 (4.6--5.5 directions) shares no initialization with the scored adapter. It keeps 0.56--0.66 over the held interval against 0.15--0.23 at the switch, 2.53--2.91 times the loss per unit of length (Table~\ref{tab:lenbody}) and 2.18--2.50 per unit of marker drop.

\paragraph{Floors and regression.}
Without the half-data reference, we score a region from two $\lambda=0$ seeds above a floor. With the $\lambda=1$ coverage as floor, it keeps 0.45--0.53 over pair A's held interval and 0.07--0.09 over its switch, and with chance as floor 0.48--0.57 and 0.15--0.24. Pair A's switch thus loses 1.93--3.07 times as much per unit of length (Table~\ref{tab:lenbody}). Pair B separates only with the $\lambda=1$ floor. Pair C's first step does not separate with either floor, and its second does. The ratio is unstable where solutions barely shorten, as in pair A after the switch. A regression of the loss on log-length drop and a switch indicator gives a coefficient of 0.46. Its cell-bootstrap interval spans zero, with an unstable lower end ($-0.11$ to $-1.53$), because in pairs B and C length and class are confounded. Per unit of log-length drop, $R$ falls by 0.36 in held intervals and 0.52 at switches, so $R$ partly follows length.

\paragraph{Pre-specified and post hoc analyses.}
The overlap ratio $R$ and its permutation test, the eigenvalue cut, the switch definition, the edits and their controls except random and post-switch removal, the whole-response marker count and the logit-lens score were fixed before the edit results were seen. The interval classes, half-data reference, floors, symmetric $R$, bootstrap intervals, cut-0.75 cross-seed build, per-marker normalization, two-seed region, structure measures, $\kappa$, pair-level intervals, marker-to-weight ratios, loop count, $\lambda=0$-region rates, overlap normalization and $\lambda=1$-versus-untrained comparison came later, and the early-marker count after grafts looped.

\section{Results per Cell}
\label{app:outputs}
\label{app:edits}
\label{app:general}

\begin{figure*}[!tp]
\centering
\includegraphics[width=\textwidth]{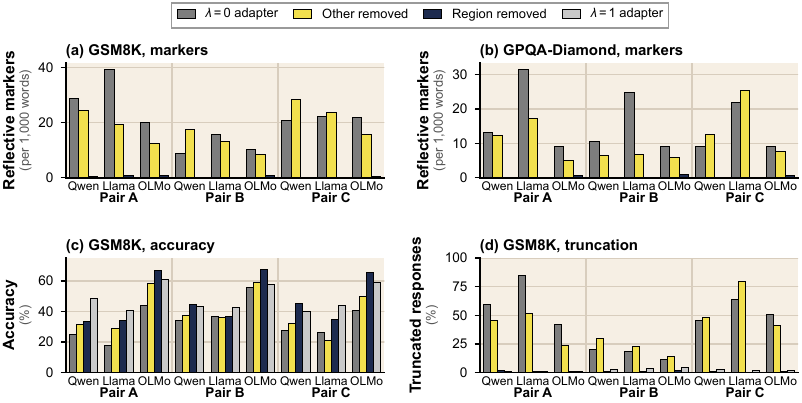}
\caption{Bars give each cell's seed mean for the MATH-trained $\lambda=0$ adapter, with other directions or the shared region removed, and the $\lambda=1$ adapter, as markers on (a) GSM8K and (b) GPQA-Diamond and (c) accuracy and (d) truncation on GSM8K. Removing the region leaves 0.2--9\% of the markers on GSM8K and 0.1--10\% on GPQA-Diamond and cuts GSM8K truncation from 11--85\% to 0.4--1.9\%. Nine-cell means are in Table~\ref{tab:general}.}
\label{fig:ood}
\end{figure*}

\paragraph{Student outputs.}
The $\lambda=1$ student falls below untrained in six of nine cells (Tables~\ref{tab:2} and~\ref{tab:untrained}), also on completed responses (by 1.2--8.5 pp). Its higher truncation (five cells) thus explains only part of the drop. Accuracy peaks at $\lambda=0.8$ in six of nine cells (tied with $\lambda=1$ in B/OLMo) and at $\lambda=1$ in the other three (Table~\ref{tab:acc}). Table~\ref{tab:examples} gives outputs chosen by a fixed rule. Figure~\ref{fig:ood} gives the benchmark part of Table~\ref{tab:general} per cell. The two-seed region holds 43--67\% of a held-out $\lambda=0$ adapter's energy and 1--10\% of the $\lambda=1$ adapter's, against 0.2--2.3\% by chance.

\paragraph{Controls and grafts.}
Several cells lengthen and stop finishing before markers rise (Table~\ref{tab:9}). The region thus turns on the long, open-ended mode but not the ability to close it. Removal barely changes the $\lambda=1$ adapter (whole-response markers 0.07--5.14 to 0.03--4.03, length 0.62--1.17$\times$). At $\beta=4$ and 8 (seed 0) every graft degenerates into repeated strings. Region grafts then finish at most 1\% of responses and other grafts at most 49\%, and all reach at most 8.0\% accuracy, so two doses are usable. $\kappa$ pairs the 1,200 responses with the untrained student's on the same problem, seed and sample, and its self-ceiling pairs the untrained student with itself across seeds. The $\beta=2$ graft recovers 97\%, 163\% and 37\% of the early verification gap in pair A's Qwen, Llama and OLMo, 43\%, 194\% and 14\% in pair C and $-15$ to 11\% in pair B. It finishes 22\% and 19\% of first responses in A/Qwen and A/Llama (recap 0.99 and 0.85).

\paragraph{Logit lens.}
A logit lens maps each region direction of \texttt{o\_proj} and \texttt{down\_proj} in the last third of the layers through the final norm and the unembedding. The pre-specified score is the mean vocabulary percentile of single-token cue words. It is 64.8 for the region and 59.8 for random directions (above random in all nine cells). It is 64.7 for the post-switch region, which the region beats in only five cells, so the score does not separate the two. An exploratory count chosen after inspecting tokens gives 1.1--4.2, 0.3--1.6 and 0.04--0.07 cue tokens among a direction's top 100 for the region, the post-switch region and random directions.

\paragraph{Seeds and intervals.}
Removal beats the cut in accuracy for 25 of 27 seeds and top removal for 19 of 27. All 27 grafts at $\beta=2$ reach at least 5.0 early markers, above every $\lambda=1$ cell mean (at most 1.5). Paired problem bootstraps with one shared draw give +3.7 pp [2.6, 4.8] over top removal and +7.9 [6.2, 9.6] over other directions (the cut is in \S\ref{sec:remove}). These intervals, like those of Table~\ref{tab:stats} with cells fixed, condition on the nine cells. With cells as units they are [$-0.2$, 7.7] and [3.5, 12.3], and with teachers as units [$-1.2$, 8.7] and [$-1.6$, 17.4]. Per-cell intervals lie above zero in 7, 3 and 7 cells for the cut, top removal and other directions (5, 3 and 6 with seeds resampled). With three seeds per cell, they reflect little seed variation.

\setcounter{bottomnumber}{4}
\renewcommand{\bottomfraction}{0.95}
\renewcommand{\textfraction}{0.03}
\setcounter{topnumber}{5}
\renewcommand{\topfraction}{0.98}
\renewcommand{\floatpagefraction}{0.5}
\setcounter{dbltopnumber}{3}
\renewcommand{\dbltopfraction}{0.98}
\renewcommand{\dblfloatpagefraction}{0.5}
\setlength{\floatsep}{5pt plus 2pt minus 2pt}
\makeatletter\setlength{\@fptop}{0pt plus 1fil}\setlength{\@fpsep}{8pt plus 2pt}\setlength{\@fpbot}{0pt plus 1fil}\setlength{\@dblfptop}{0pt}\setlength{\@dblfpsep}{8pt plus 2pt}\setlength{\@dblfpbot}{0pt plus 1fil}\makeatother
\setlength{\dblfloatsep}{8pt plus 2pt minus 3pt}

\begin{table}[!hbp]
\centering
\small
\setlength{\tabcolsep}{4pt}
\begin{adjustbox}{max width=\columnwidth}\begin{tabular}{lrrrrrrr}
\toprule
 & & \multicolumn{3}{c}{Energy in region (\%)} & \multicolumn{3}{c}{$\lambda=0$ by depth (\%)} \\
\cmidrule(lr){3-5}\cmidrule(lr){6-8}
Cell & Dims & $\lambda=0$ & $\lambda=1$ & Chance & Early & Middle & Late \\
\midrule
A/Qwen & 6.5 & 47 & 1 & 2.4 & 35 & 42 & 55 \\
A/Llama & 6.6 & 55 & 4 & 0.6 & 46 & 55 & 62 \\
A/OLMo & 6.4 & 64 & 6 & 0.3 & 57 & 64 & 68 \\
\midrule
B/Qwen & 8.0 & 56 & 10 & 2.7 & 45 & 53 & 64 \\
B/Llama & 7.6 & 64 & 12 & 0.7 & 56 & 64 & 70 \\
B/OLMo & 7.8 & 72 & 15 & 0.3 & 66 & 72 & 76 \\
\midrule
C/Qwen & 7.6 & 52 & 2 & 2.6 & 44 & 47 & 60 \\
C/Llama & 7.6 & 59 & 2 & 0.7 & 51 & 58 & 65 \\
C/OLMo & 7.6 & 68 & 11 & 0.3 & 62 & 69 & 71 \\
\bottomrule
\end{tabular}
\end{adjustbox}
\caption{Region built from all pre-switch adapters (unweighted, seed means). Columns give its dimensions, its share of the $\lambda=0$ and $\lambda=1$ update energy against a random subspace of its size, and its share of the $\lambda=0$ energy by third of the layers. Each module type holds 4.2--10.5 region directions.}
\label{tab:anatomy}
\end{table}

\begin{table}[!tbp]
\centering
\small
\begin{adjustbox}{max width=\columnwidth}\begin{tabular}{lrrrrrrrrr}
\toprule
 & \multicolumn{3}{c}{Pair A} & \multicolumn{3}{c}{Pair B} & \multicolumn{3}{c}{Pair C} \\
\cmidrule(lr){2-4}\cmidrule(lr){5-7}\cmidrule(lr){8-10}
 & Qwen & Llama & OLMo & Qwen & Llama & OLMo & Qwen & Llama & OLMo \\
\midrule
\rowcolor{band}\multicolumn{10}{l}{\emph{Whole-response markers per 1,000 words}} \\
$\lambda=0$ & 21.8 & 30.7 & 15.0 & 19.2 & 44.4 & 16.8 & 16.6 & 19.1 & 14.4 \\
$\gamma=1$ & 21.2 & 20.9 & 10.1 & 10.8 & 18.8 & 10.8 & 21.7 & 19.1 & 11.4 \\
$\gamma=4$ & 19.4 & 16.2 & 8.7 & 11.3 & 11.6 & 9.9 & 25.6 & 18.3 & 12.5 \\
All in region & 19.1 & 13.7 & 7.4 & 11.1 & 9.9 & 9.4 & 27.3 & 19.6 & 12.1 \\
\midrule
\rowcolor{band}\multicolumn{10}{l}{\emph{Early markers per 1,000 words}} \\
$\lambda=0$ & 10.2 & 12.4 & 10.1 & 15.7 & 16.2 & 15.1 & 10.2 & 12.5 & 10.1 \\
$\gamma=1$ & 13.0 & 13.5 & 10.2 & 17.1 & 17.6 & 16.0 & 13.8 & 15.8 & 11.3 \\
$\gamma=4$ & 14.8 & 13.3 & 9.8 & 17.6 & 16.6 & 16.1 & 16.2 & 18.4 & 12.1 \\
All in region & 14.9 & 13.3 & 8.9 & 16.3 & 14.1 & 15.1 & 18.4 & 19.1 & 11.3 \\
\midrule
\rowcolor{band}\multicolumn{10}{l}{\emph{Region weight relative to $\lambda=0$ ($\times$)}} \\
$\gamma=1$ & 1.35 & 1.29 & 1.22 & 1.25 & 1.20 & 1.15 & 1.28 & 1.23 & 1.17 \\
$\gamma=4$ & 1.56 & 1.45 & 1.33 & 1.37 & 1.29 & 1.21 & 1.42 & 1.34 & 1.24 \\
All in region & 1.62 & 1.49 & 1.35 & 1.40 & 1.31 & 1.22 & 1.46 & 1.36 & 1.26 \\
\bottomrule
\end{tabular}
\end{adjustbox}
\caption{The region raised by $\gamma$, or the whole update moved into it (All), at a fixed norm (seed means). Early markers change 0.96--1.58$\times$ at $\gamma \le 4$ and 0.87--1.79$\times$ with all in region, while nine-cell accuracy falls from 13.9\% to 10.1--11.5\% and completion from 37\% to 28--31\%.}
\label{tab:10}
\end{table}

\begin{table}[!tbp]
\centering
\small
\setlength{\tabcolsep}{4pt}
\begin{adjustbox}{max width=\columnwidth}\begin{tabular}{lrrrrrrr}
\toprule
 & \multicolumn{3}{c}{Pair A} & \multicolumn{2}{c}{Pair B} & \multicolumn{2}{c}{Pair C} \\
\cmidrule(lr){2-4}\cmidrule(lr){5-6}\cmidrule(lr){7-8}
Measure / $\lambda$ & 0 & 0.2 & 0.4 & 0 & 0.4 & 0 & 0.4 \\
\midrule
Markers & 15.5 & 5.0 & 1.0 & 13.3 & 10.9 & 14.2 & 11.2 \\
Words & 1,348 & 679 & 289 & 1,737 & 1,146 & 1,209 & 794 \\
Tag (\%) & 100 & 96 & 0 & 100 & 100 & 100 & 100 \\
Recap (\%) & 100 & 96 & 0 & 100 & 100 & 100 & 100 \\
Opener (\%) & 100 & 85 & 0 & 65 & 58 & 100 & 100 \\
Sent. (\%) & 28 & 18 & 3 & 19 & 21 & 28 & 24 \\
Verif. & 1.65 & 1.60 & 0.52 & 1.67 & 2.31 & 1.68 & 1.81 \\
Check (\%) & 100 & 80 & 23 & 100 & 100 & 100 & 96 \\
Hesit. & 8.48 & 2.34 & 0.03 & 9.03 & 6.31 & 7.75 & 5.66 \\
Altern. & 1.64 & 0.27 & 0.02 & 0.17 & 0.36 & 1.46 & 0.96 \\
Attempts & 4.65 & 1.39 & 1.01 & 2.60 & 2.13 & 3.92 & 2.19 \\
\bottomrule
\end{tabular}
\end{adjustbox}
\caption{Teacher training solutions at the ends of each held interval and, for pair A, after its switch: markers, words, the think tag, the recap after it, a monologue opener, the share of monologue sentences, verification, hesitation and alternative markers per 1,000 words, the share with any check, and attempts per solution.}
\label{tab:structure}
\end{table}

\begin{table}[!tbp]
\centering
\small
\setlength{\tabcolsep}{3.5pt}
\begin{adjustbox}{max width=\columnwidth}\begin{tabular}{lrrrrrrrr}
\toprule
 & \multicolumn{2}{c}{Acc.\ (\%)} & \multicolumn{2}{c}{Tokens} & & \multicolumn{3}{c}{$\kappa$ with untrained} \\
\cmidrule(lr){2-3}\cmidrule(lr){4-5}\cmidrule(lr){7-9}
Cell & Rem. & Untr. & Rem. & Untr. & Okay (\%) & Rem. & $\lambda=1$ & Self \\
\midrule
A/Qwen & 20.2 & \textbf{35.1} & 1,439 & 940 & 87 & 0.31 & 0.49 & 0.57 \\
A/Llama & 22.1 & \textbf{23.3} & 1,659 & 863 & 25 & 0.34 & 0.33 & 0.46 \\
A/OLMo & 19.8 & \textbf{22.2} & 732 & 637 & 0 & 0.52 & 0.49 & 0.57 \\
\midrule
B/Qwen & \textbf{35.8} & 35.1 & 966 & 940 & 1 & 0.50 & 0.44 & 0.57 \\
B/Llama & \textbf{23.7} & 23.3 & 1,141 & 863 & 5 & 0.37 & 0.35 & 0.46 \\
B/OLMo & 21.4 & \textbf{22.2} & 760 & 637 & 0 & 0.52 & 0.38 & 0.57 \\
\midrule
C/Qwen & 34.5 & \textbf{35.1} & 1,025 & 940 & 2 & 0.48 & 0.42 & 0.57 \\
C/Llama & \textbf{23.6} & 23.3 & 1,214 & 863 & 0 & 0.43 & 0.39 & 0.46 \\
C/OLMo & 20.9 & \textbf{22.2} & 744 & 637 & 0 & 0.54 & 0.35 & 0.57 \\
\bottomrule
\end{tabular}
\end{adjustbox}
\caption{Region removal (Rem.)\ against the untrained student (Untr.)\ on MATH-500 (seed means). Okay is the share of first responses opening with ``Okay''. $\kappa$ is Cohen's $\kappa$ of response-level correctness with the untrained student, and Self pairs the untrained student with itself across seeds. In pairs B and C removal agrees with the untrained student at least as well as the $\lambda=1$ adapter does. Bold marks the higher accuracy.}
\label{tab:untrained}
\end{table}

\begin{table}[!tbp]
\centering
\small
\setlength{\tabcolsep}{4pt}
\begin{adjustbox}{max width=\columnwidth}\begin{tabular}{lrrrrrr}
\toprule
 & & \multicolumn{2}{c}{Ratio reached} & & \multicolumn{2}{c}{Dims removed} \\
\cmidrule(lr){3-4}\cmidrule(lr){6-7}
Cell & Region & $\lVert B \rVert^2$ & $\lVert \Delta W \rVert^2$ & Short (\%) & Region & Other \\
\midrule
A/Qwen & 0.34 & 1.06 & 0.97 & 19 & 7.4 & 7.8 \\
A/Llama & 0.39 & 1.03 & 0.89 & 13 & 7.3 & 8.6 \\
A/OLMo & 0.48 & 0.92 & 0.73 & 50 & 7.2 & 12.2 \\
\midrule
B/Qwen & 0.45 & 0.94 & 0.77 & 44 & 8.4 & 11.7 \\
B/Llama & 0.50 & 0.85 & 0.63 & 55 & 8.0 & 13.3 \\
B/OLMo & 0.58 & 0.68 & 0.47 & 79 & 8.3 & 15.0 \\
\midrule
C/Qwen & 0.44 & 0.96 & 0.89 & 46 & 7.9 & 12.0 \\
C/Llama & 0.49 & 0.90 & 0.76 & 57 & 7.9 & 13.6 \\
C/OLMo & 0.59 & 0.68 & 0.56 & 80 & 7.9 & 15.1 \\
\bottomrule
\end{tabular}
\end{adjustbox}
\caption{Energy audit of region removal (seed means). Region is the region's share of $\lVert B \rVert^2$. The next two columns give the share of it that the other-directions control reaches in $\lVert B \rVert^2$ (matched per module) and in $\lVert \Delta W \rVert^2$. Short is the share of modules where the control falls short, and the last two columns give the directions removed per module.}
\label{tab:energy}
\end{table}

\begin{table}[!tbp]
\centering
\small
\setlength{\tabcolsep}{4pt}
\begin{adjustbox}{max width=\columnwidth}\begin{tabular}{lllrrrrrrr}
\toprule
 & & & & & \multicolumn{3}{c}{Kept} & \multicolumn{2}{c}{$\Delta$Coverage} \\
\cmidrule(lr){6-8}\cmidrule(lr){9-10}
Pair & $\lambda$ & Class & $\Delta\log L$ & Think & Qwen & Llama & OLMo & LOO & $\lambda$-out \\
\midrule
A & 0--0.2 & Held & 0.67 & 91 & 0.50 & 0.57 & 0.58 & $-$0.02 & $-$0.01 \\
 & 0.2--0.4 & Switch & 0.80 & 0 & 0.12 & 0.19 & 0.18 & $-$0.33 & $-$0.22 \\
 & 0.4--0.6 & After & 0.10 & 0 & 0.29 & 0.68 & 0.62 & $-$0.02 & $-$0.02 \\
 & 0.6--0.8 & After & 0.06 & 0 & 0.35 & 0.79 & 0.75 & $-$0.01 & $-$0.01 \\
 & 0.8--1.0 & After & 0.04 & 0 & 0.36 & 0.81 & 0.77 & 0.00 & 0.00 \\
\midrule
B & 0--0.2 & Held & 0.21 & 83 & 0.88 & 0.92 & 0.96 & 0.03 & 0.06 \\
 & 0.2--0.4 & Held & 0.19 & 91 & 0.83 & 0.87 & 0.91 & $-$0.04 & $-$0.09 \\
 & 0.4--0.6 & Switch & 0.68 & 1 & 0.40 & 0.41 & 0.45 & $-$0.27 & $-$0.28 \\
 & 0.6--0.8 & After & 0.51 & 0 & 0.43 & 0.48 & 0.50 & $-$0.11 & $-$0.10 \\
 & 0.8--1.0 & After & 0.01 & 0 & 0.80 & 0.89 & 0.90 & $-$0.01 & $-$0.01 \\
\midrule
C & 0--0.2 & Held & 0.22 & 90 & 0.83 & 0.89 & 0.92 & 0.02 & 0.05 \\
 & 0.2--0.4 & Held & 0.20 & 89 & 0.78 & 0.86 & 0.89 & $-$0.04 & $-$0.08 \\
 & 0.4--0.6 & Switch & 0.60 & 27 & 0.44 & 0.52 & 0.56 & $-$0.21 & $-$0.22 \\
 & 0.6--0.8 & Switch & 0.69 & 0 & 0.16 & 0.13 & 0.28 & $-$0.24 & $-$0.22 \\
 & 0.8--1.0 & After & $-$0.06 & 0 & 0.43 & 0.30 & 0.66 & $-$0.01 & $-$0.01 \\
\bottomrule
\end{tabular}
\end{adjustbox}
\caption{Intervals classified by the teacher as Held, Switch or After. Columns give the log training-length drop and the teacher's think rate at the end (83\%, 73\% and 90\% at $\lambda=0$). Kept is the share of the region built at the interval's start that the end adapters keep relative to same-$\lambda$ half-data adapters (post hoc). The last two columns give the change in leave-one-out coverage (in-sample for pair A's held interval) and in coverage of the region rebuilt without the scored $\lambda$ (pre-specified, symmetric by construction there), as student means. After-switch shares score post-switch regions, so they are no floor.}
\label{tab:length}
\end{table}

\begin{table}[!tbp]
\centering
\small
\setlength{\tabcolsep}{4pt}
\begin{adjustbox}{max width=\columnwidth}\begin{tabular}{lrrrrrrr}
\toprule
 & \multicolumn{2}{c}{Tokens} & \multicolumn{2}{c}{Trunc.\ (\%)} & \multicolumn{2}{c}{$\Delta$Acc.\ (pp)} & \\
\cmidrule(lr){2-3}\cmidrule(lr){4-5}\cmidrule(lr){6-7}
Cell & $\lambda=0$ & $\lambda=1$ & $\lambda=0$ & $\lambda=1$ & All & Compl. & $r$ \\
\midrule
A/Qwen & 5,916 & 811 & 70 & 4 & +19.2 & $-$4.4 & 1.00 \\
A/Llama & 6,969 & 944 & 87 & 5 & +18.1 & $-$14.3 & 1.00 \\
A/OLMo & 3,015 & 683 & 61 & 5 & +9.6 & +1.4 & 1.00 \\
\midrule
B/Qwen & 5,226 & 2,302 & 55 & 22 & +8.8 & $-$2.8 & 0.94 \\
B/Llama & 5,095 & 1,765 & 52 & 14 & +6.5 & $-$6.2 & 0.82 \\
B/OLMo & 2,827 & 1,207 & 51 & 16 & +5.7 & $-$0.2 & 0.89 \\
\midrule
C/Qwen & 5,231 & 1,399 & 59 & 13 & +8.7 & $-$5.8 & 0.97 \\
C/Llama & 6,149 & 939 & 73 & 6 & +15.3 & $-$10.2 & 0.94 \\
C/OLMo & 2,949 & 910 & 58 & 14 & +1.7 & $-$4.6 & 0.99 \\
\bottomrule
\end{tabular}
\end{adjustbox}

\vspace{5pt}
\begin{adjustbox}{max width=\columnwidth}\begin{tabular}{lrrrrrr}
\toprule
Cell & $\lambda=0$ & 0.2 & 0.4 & 0.6 & 0.8 & 1.0 \\
\midrule
A/Qwen & 15.2 & 25.8 & 33.8 & \underline{36.7} & \textbf{37.7} & 34.4 \\
A/Llama & 11.8 & 24.2 & 28.5 & \underline{29.0} & 28.6 & \textbf{29.8} \\
A/OLMo & 10.2 & 14.4 & \underline{19.3} & 17.1 & 18.3 & \textbf{19.8} \\
\midrule
B/Qwen & 17.5 & 17.7 & 18.3 & \underline{26.5} & \textbf{29.2} & 26.2 \\
B/Llama & 17.1 & 19.3 & 16.1 & \underline{24.2} & \textbf{26.9} & 23.6 \\
B/OLMo & 10.5 & 11.8 & 11.8 & \underline{15.5} & \textbf{16.2} & \textbf{16.2} \\
\midrule
C/Qwen & 16.2 & 18.1 & 19.8 & \underline{26.4} & \textbf{27.1} & 24.9 \\
C/Llama & 15.0 & 17.3 & 19.1 & 25.2 & \underline{28.7} & \textbf{30.3} \\
C/OLMo & 11.7 & 11.3 & 11.4 & \underline{14.5} & \textbf{15.6} & 13.3 \\
\bottomrule
\end{tabular}
\end{adjustbox}
\caption{MATH-500 results of the students (seed means): tokens and truncation at $\lambda=0$ and 1, the accuracy change over all and completed responses, $r$ of the student's markers with the teacher's over the six $\lambda$ (0.86 over all 54 points), and accuracy (\%) at every $\lambda$ (bottom). Bold and underline mark each cell's best and second-best accuracy over $\lambda$.}
\label{tab:2}
\label{tab:acc}
\end{table}

\begin{table*}[!tp]
\centering
\small
\setlength{\tabcolsep}{4pt}
{\renewcommand{\arraystretch}{0.92}\begin{adjustbox}{max width=\textwidth}\begin{tabular}{lrrrrrrrrr}
\toprule
 & \multicolumn{3}{c}{Pair A} & \multicolumn{3}{c}{Pair B} & \multicolumn{3}{c}{Pair C} \\
\cmidrule(lr){2-4}\cmidrule(lr){5-7}\cmidrule(lr){8-10}
 & Qwen & Llama & OLMo & Qwen & Llama & OLMo & Qwen & Llama & OLMo \\
\midrule
\rowcolor{band}\multicolumn{10}{l}{\emph{Whole-response markers per 1,000 words}} \\
$\lambda=0$ & 21.8 & 30.7 & 15.0 & 19.2 & 44.4 & 16.8 & 16.6 & 19.1 & 14.4 \\
Region removed & 0.37 & 0.18 & 1.16 & 0.15 & 0.07 & 1.45 & 0.38 & 0.10 & 1.43 \\
Length-matched cut & 13.8 & 19.4 & 11.2 & 14.0 & 22.1 & 14.3 & 11.7 & 16.3 & 11.9 \\
Other removed & 19.1 & 17.9 & 8.8 & 15.2 & 12.8 & 7.7 & 24.3 & 22.2 & 12.9 \\
Top removed & 6.03 & 10.58 & 3.29 & 1.15 & 0.58 & 1.71 & 2.95 & 0.81 & 1.47 \\
Scaled down & 13.0 & 27.6 & 10.3 & 4.6 & 6.8 & 5.4 & 8.4 & 13.8 & 8.8 \\
Random removed & 9.02 & 9.64 & 2.27 & 0.48 & 0.69 & 1.53 & 5.09 & 5.39 & 2.00 \\
Post-switch removed & 26.8 & 46.2 & 27.0 & 1.7 & 6.1 & 17.5 & 2.2 & 4.3 & 19.6 \\
\midrule
\rowcolor{band}\multicolumn{10}{l}{\emph{Early markers per 1,000 words}} \\
$\lambda=0$ & 10.16 & 12.36 & 10.14 & 15.72 & 16.17 & 15.14 & 10.24 & 12.48 & 10.07 \\
Region removed & 0.43 & 0.07 & 1.10 & 0.23 & 0.08 & 1.29 & 0.33 & 0.07 & 1.33 \\
Length-matched cut & 10.41 & 12.79 & 9.94 & 14.00 & 16.45 & 14.14 & 9.79 & 12.56 & 10.47 \\
Other removed & 10.22 & 10.23 & 7.41 & 12.39 & 9.80 & 10.11 & 12.26 & 14.82 & 9.63 \\
Top removed & 3.72 & 3.82 & 1.92 & 1.24 & 0.25 & 1.47 & 1.67 & 0.41 & 1.22 \\
Scaled down & 7.48 & 8.74 & 7.40 & 6.44 & 7.65 & 3.79 & 6.76 & 7.92 & 5.46 \\
Random removed & 4.15 & 4.25 & 1.68 & 0.55 & 0.25 & 1.34 & 2.93 & 1.77 & 1.20 \\
Post-switch removed & 10.74 & 18.03 & 12.84 & 4.19 & 8.85 & 6.50 & 2.28 & 3.88 & 8.52 \\
\midrule
\rowcolor{band}\multicolumn{10}{l}{\emph{MATH-500 accuracy (\%)}} \\
$\lambda=0$ & 15.2 & 11.8 & 10.2 & 17.5 & 17.1 & 10.5 & 16.2 & 15.0 & 11.7 \\
Region removed & \textbf{20.2} & \textbf{22.1} & 19.8 & \textbf{35.8} & \underline{23.7} & \textbf{21.4} & \textbf{34.5} & \textbf{23.6} & 20.9 \\
Length-matched cut & 18.8 & \underline{21.0} & 11.2 & 14.7 & 20.1 & 11.2 & 17.4 & 19.5 & 10.5 \\
Other removed & 16.9 & 19.4 & 14.5 & 19.2 & 21.4 & 14.5 & 16.8 & 14.5 & 13.7 \\
Top removed & 18.8 & 17.2 & \textbf{21.8} & 25.3 & 21.9 & 20.4 & 20.9 & 21.2 & \underline{21.0} \\
Scaled down & 18.3 & 16.8 & 17.3 & 22.2 & 21.9 & 19.2 & 20.7 & \underline{21.8} & 20.2 \\
Random removed & \underline{19.0} & 19.6 & \underline{21.2} & \underline{30.8} & \textbf{26.2} & \underline{20.6} & \underline{23.6} & 20.8 & 20.7 \\
Post-switch removed & 15.1 & 6.7 & 12.8 & 16.3 & 18.3 & 16.2 & 22.9 & 15.6 & \textbf{21.1} \\
\bottomrule
\end{tabular}
\end{adjustbox}}
\caption{The $\lambda=0$ adapter with the region removed, cut at the same length, with energy removed from other directions (at most the region's) or the top singular components (matched), or scaled down (seed means). Nine-cell mean accuracy is 13.9, 24.7, 16.0, 16.8, 21.0, 19.8, 22.5 and 16.1\% in row order. Random removal takes randomly rotated update directions per module until it removes the region's $\lVert \Delta W \rVert^2$, reaching 1.06--1.12 of it over cells and seeds (one rotation per seed) with 6.9--10.7 directions per module. The same-size post-switch region removes 0.06--0.33 of that energy. Both were run after the edit results. Other-direction removal takes 0.46--0.98 of the region's $\lVert \Delta W \rVert^2$, about half in B/OLMo and C/OLMo (Table~\ref{tab:energy}). B/Llama's $\lambda=0$ marker mean rests on seeds of 66.8, 15.9 and 50.4. Bold and underline mark the best and second-best accuracy per cell.}
\label{tab:6}
\label{tab:7}

\vspace{3pt}
\begin{minipage}[t]{0.485\textwidth}
\vspace{0pt}
\centering
\footnotesize
\begin{adjustbox}{max width=\linewidth}\begin{tabular}{llrrrrr}
\toprule
Signal & Pair & Qwen & Llama & OLMo & Mean & $p$ \\
\midrule
Think rate & A & 0.53 & 0.84 & 0.80 & 0.72 & 0.200 \\
 & B & 0.94 & 0.95 & 0.94 & 0.94 & 0.050 \\
 & C & 0.91 & 0.83 & 0.81 & 0.85 & 0.067 \\
 & Pooled & & & & 0.84 & 0.0035 \\
\midrule
Markers & A & 0.81 & 0.55 & 0.61 & 0.66 & 0.150 \\
 & B & 0.94 & 0.95 & 0.94 & 0.94 & 0.075 \\
 & C & 0.93 & 0.88 & 0.87 & 0.89 & 0.075 \\
 & Pooled & & & & 0.83 & 0.0032 \\
\midrule
Length & A & 0.94 & 0.95 & 0.97 & 0.95 & 0.050 \\
 & B & 0.89 & 0.88 & 0.90 & 0.89 & 0.017 \\
 & C & 0.86 & 0.90 & 0.90 & 0.89 & 0.058 \\
 & Pooled & & & & 0.91 & 0.0002 \\
\midrule
Completed & A & 0.62 & 0.90 & 0.86 & 0.79 & 0.050 \\
 & B & 0.94 & 0.95 & 0.94 & 0.94 & 0.017 \\
 & C & 0.95 & 0.89 & 0.87 & 0.90 & 0.100 \\
 & Pooled & & & & 0.88 & 0.0011 \\
\midrule
Training & A & 0.63 & 0.90 & 0.86 & 0.80 & 0.050 \\
 & B & 0.94 & 0.95 & 0.94 & 0.94 & 0.050 \\
 & C & 0.96 & 0.90 & 0.89 & 0.92 & 0.100 \\
 & Pooled & & & & 0.89 & 0.0006 \\
\bottomrule
\end{tabular}
\end{adjustbox}
\caption{Correlation $r$ of the per-interval drops in $R$ with those in a teacher signal, and the exact permutation $p$. Length is the log training-solution length, and Completed and Training are the think rate over completed solutions or the training problems. No single pair reaches $p<0.05$ with the think rate or markers, and training length gives the smallest pooled $p$ (\S\ref{sec:directions}).}
\label{tab:4}
\end{minipage}\hfill
\begin{minipage}[t]{0.485\textwidth}
\vspace{0pt}
\centering
\footnotesize
\setlength{\tabcolsep}{4pt}
\begin{adjustbox}{max width=\linewidth}\setlength{\tabcolsep}{2.6pt}
\begin{tabular}{lrrrrrr}
\toprule
Measure & $\lambda=0$ & Rem. & Other & Graft & $\lambda=1$ & Untr. \\
\midrule
Hesitation & 9.85 & 0.03 & 7.84 & 15.64 & 0.20 & 0.01 \\
Verification & 2.17 & 0.51 & 2.19 & 1.05 & 0.24 & 0.28 \\
Attempts & 1.15 & 1.00 & 1.27 & 1.22 & 1.01 & 1.00 \\
Opener & 0.85 & 0.15 & 0.85 & 0.63 & 0.00 & 0.00 \\
Mono.\ sent. & 0.42 & 0.03 & 0.36 & 0.41 & 0.04 & 0.02 \\
Recap (compl.) & 0.79 & 0.03 & 0.77 & -- & 0.04 & 0.02 \\
\bottomrule
\end{tabular}
\end{adjustbox}

\begin{adjustbox}{max width=\linewidth}\setlength{\tabcolsep}{4pt}
\begin{tabular}{lrrr}
\toprule
SAE thinking features & Pair A & Pair B & Pair C \\
\midrule
$\lambda=0$ & 0.89 & 0.68 & 0.93 \\
$\lambda=1$ & 0.093 & 0.199 & 0.089 \\
Region removed & 0.21 & 0.14 & 0.13 \\
Other removed & 0.94 & 0.62 & 0.91 \\
Graft region ($\beta=2$) & 0.46 & 0.19 & 0.64 \\
Graft other ($\beta=2$) & 0.13 & 0.19 & 0.10 \\
Graft whole update ($\beta=2$) & 0.47 & 0.19 & 0.67 \\
All in region & 1.02 & 0.53 & 0.92 \\
$\lambda=0$, markers deleted & 0.79 & 0.64 & 0.86 \\
\midrule
$r$ with markers & 0.88 & $-0.08$ & 0.92 \\
\bottomrule
\end{tabular}
\end{adjustbox}
\caption{Top: nine-cell student structure means in the first 300 words (recap on completed responses; the graft's only in A/Qwen, 0.99, and A/Llama, 0.85). Bottom: the Llama student's SAE thinking features on its own answers (1 on $\lambda=0$ teacher solutions) and their $r$ with markers over 57 non-independent points. It is only an auxiliary read-out (Appendix~\ref{app:details}).}
\label{tab:studstruct}
\label{tab:11}
\end{minipage}
\end{table*}

\begin{table*}[!tp]
\centering
\setlength{\tabcolsep}{3pt}\footnotesize
\begin{tabular}{@{}p{0.12\textwidth}p{0.64\textwidth}rrp{0.08\textwidth}@{}}
\toprule
Output & First 30 and last 40 words & Tokens & Markers & End \\
\midrule
$\lambda=0$ student & Alright, so I need to figure out how many positive whole-number divisors 196 has. \textbf{Hmm}, okay, let me think about this step by step. I remember that a divisor of \ldots{} 16 = 12.25, 196 / 24.5 = 8.0625, 196 / 49 = 4, 196 / 98 = 2, 196 / 196 = 1, which is 1 more. So, that's 196 / 1 = 2, 196 / 2 = 98, 196 & 7,680 & 2.4 & truncated, loop \\
\addlinespace
Region removed & Alright, let's think about this step by step. We're looking for the number of positive whole-number divisors of 196. 1. Understanding the Problem: - We need to find the number \ldots{} . - For [math] , the exponents are [math] and [math] . - Applying the formula, we get: [math] 4. Conclusion: - The number of positive whole-number divisors of 196 is [math] . Therefore, the final answer is [math] . & 458 & 0.0 & answered (correct) \\
\addlinespace
Length-matched cut & Alright, so I need to figure out how many positive whole-number divisors 196 has. \textbf{Hmm}, okay, let me think about this step by step. I remember that a divisor of \ldots{} 6.1875, not an integer. 33: 196 / 33 approx. 5.838, not an integer. 34: 196 / 34 = 5.714, not an integer. 35: 196 / 35 approx. 5.514, not an integer. 36: 196 / 36 approx. 5.472 Final Answer [math] & 1,502 & 8.1 & cut, then answered (correct) \\
\addlinespace
Graft, $\beta=2$ & Alright, so I need to find how many positive whole-number divisors 196 has. \textbf{Hmm}, okay, so dividing 196 by some numbers and seeing how many times it divides evenly. Let \ldots{} So, 196 is 2 squared times 49 squared, which is 4 times 2001. So, 196 is 2 squared times 49 squared, which is 4 times 2001. So, 196 is 2 squared times 49 squared, which is 4 times 2001. So & 7,680 & 1.1 & truncated, loop \\
\addlinespace
$\lambda=1$ student & To determine the number of positive whole-number divisors of 196, we first need to find its prime factorization. We start by dividing 196 by the smallest prime number, which is \ldots{} 2 + 2 = 4. Therefore, the number of positive whole-number divisors is: [math] The positive whole-number divisors of 196 are 1, 2, 4, 7, 14, 49, and 196. Thus, the number of positive whole-number divisors of 196 is: [math] & 397 & 0.0 & answered (wrong) \\
\addlinespace
\midrule
Teacher $\lambda=0$ & Okay, so I have this vector problem here. Let me read it again and try to understand what's being asked. We have three vectors: a, b, and c. Their magnitudes \ldots{} This simplifies to: [math] Taking the magnitude of both sides: [math] We compute the magnitude squared: [math] Substituting the known values: [math] This equals [math] since [math] : [math] Solving for [math] : [math] Thus, the final answer is: [math] & 4,866 & 13.3 & training solution \\
\addlinespace
Teacher $\lambda=0.2$ & Okay, so I have this vector problem here. Let me see if I can figure it out step by step. The problem says: Let [math] , [math] , [math] be \ldots{} of both sides with [math] : [math] Since [math] , we have: [math] Thus, the equation simplifies to: [math] Since [math] , we have: [math] Using the Pythagorean identity, we find: [math] Thus, the value of [math] is [math] . & 2,824 & 1.0 & training solution \\
\addlinespace
Teacher $\lambda=0.4$ & To find [math] , we start with the given vector equation: [math] We can use the vector triple product identity, which states that for any vectors [math] , [math] , \ldots{} the identity: [math] So: [math] Taking the square root: [math] Since [math] is the angle between two vectors, it is typically taken to be between 0 and [math] , so [math] is non-negative. Therefore: [math] The final answer is: [math] & 1,386 & 0.0 & training solution \\
\addlinespace
\bottomrule
\end{tabular}

\caption{Example outputs chosen by a fixed rule. The student rows show cell A/Qwen, seed 0, on the first MATH-500 evaluation problem in file order for which the $\lambda=0$ student's first response hits the generation limit or loops, the region-removed adapter answers correctly and the $\beta=2$ graft does not finish. This is the fourth problem, ``How many positive whole-number divisors does 196 have?'' Each row gives the first 30 and the last 40 words of the first response, with math replaced by [math] and reflective markers in bold, its tokens, its markers per 1,000 words, and how it ends. A loop means that at least half of the word 10-grams in its last 300 words repeat, and the length-matched cut writes its answer after the cut. The teacher rows show pair A's training solutions at $\lambda=0$, 0.2 and 0.4 for the first training problem shared by all three, a vector problem, with tokens counted by the Qwen2.5-0.5B tokenizer. Their think tags are stripped, as in training. Removing the region leaves a short correct answer on a problem where the $\lambda=0$ student and the graft run into a loop. These two responses contain 2.4 and 1.1 markers per 1,000 words, against A/Qwen means of 21.8 and 13.0 (Tables~\ref{tab:6} and~\ref{tab:9}).}
\label{tab:examples}
\end{table*}

\begin{table}[!tbp]
\centering
\small
\setlength{\tabcolsep}{3pt}
\begin{adjustbox}{max width=\columnwidth}\begin{tabular}{lrrrr}
\toprule
Cell & Kept (\%) & Acc.\ (\%) & $-$Top (pp) & $-$Cut (pp) \\
\midrule
A/Qwen & 1.7 $\pm$ 0.4 & 20.2 $\pm$ 2.3 & 1.5 $\pm$ 3.7 & 1.5 $\pm$ 4.2 \\
A/Llama & 0.6 $\pm$ 0.2 & 22.1 $\pm$ 1.5 & 4.9 $\pm$ 2.9 & 1.1 $\pm$ 1.9 \\
A/OLMo & 7.7 $\pm$ 0.7 & 19.8 $\pm$ 0.8 & $-$2.0 $\pm$ 0.8 & 8.7 $\pm$ 1.0 \\
\midrule
B/Qwen & 0.8 $\pm$ 0.2 & 35.8 $\pm$ 0.4 & 10.5 $\pm$ 3.9 & 21.2 $\pm$ 1.2 \\
B/Llama & 0.2 $\pm$ 0.1 & 23.7 $\pm$ 2.8 & 1.8 $\pm$ 3.1 & 3.6 $\pm$ 2.1 \\
B/OLMo & 8.6 $\pm$ 0.5 & 21.4 $\pm$ 1.3 & 1.0 $\pm$ 2.0 & 10.2 $\pm$ 1.1 \\
\midrule
C/Qwen & 2.3 $\pm$ 1.5 & 34.5 $\pm$ 4.1 & 13.6 $\pm$ 3.7 & 17.1 $\pm$ 4.8 \\
C/Llama & 0.5 $\pm$ 0.6 & 23.6 $\pm$ 1.5 & 2.3 $\pm$ 1.8 & 4.1 $\pm$ 3.3 \\
C/OLMo & 9.9 $\pm$ 1.2 & 20.9 $\pm$ 0.8 & $-$0.1 $\pm$ 0.9 & 10.4 $\pm$ 0.6 \\
\midrule
Mean & 3.6 & 24.7 & 3.7 & 8.6 \\
95\% CI, cells fixed & [3.4, 3.8] & [24.0, 25.3] & [2.6, 4.8] & [6.5, 10.8] \\
95\% CI, cells as units & [0.5, 6.6] & [20.0, 29.4] & [$-$0.2, 7.7] & [3.3, 14.0] \\
Median & 1.7 & 22.1 & 1.8 & 8.7 \\
Mean w/o B/Qwen, C/Qwen & 4.2 & 21.7 & 1.3 & 5.7 \\
\bottomrule
\end{tabular}
\end{adjustbox}

\vspace{5pt}
\begin{adjustbox}{max width=\columnwidth}\begin{tabular}{lrrrr}
\toprule
Cell & Early & Done (\%) & Rec.\ $\beta=1$ & Rec.\ $\beta=2$ \\
\midrule
A/Qwen & 19.6 $\pm$ 1.5 & 23.0 $\pm$ 10.9 & 3 $\pm$ 2 & 194 $\pm$ 29 \\
A/Llama & 8.4 $\pm$ 0.7 & 17.4 $\pm$ 7.3 & 8 $\pm$ 1 & 68 $\pm$ 6 \\
A/OLMo & 5.7 $\pm$ 0.6 & 0.2 $\pm$ 0.3 & 5 $\pm$ 1 & 56 $\pm$ 9 \\
\midrule
B/Qwen & 20.4 $\pm$ 4.2 & 0.0 $\pm$ 0.0 & 71 $\pm$ 10 & 133 $\pm$ 26 \\
B/Llama & 18.7 $\pm$ 7.2 & 0.1 $\pm$ 0.1 & 46 $\pm$ 12 & 116 $\pm$ 46 \\
B/OLMo & 28.0 $\pm$ 11.2 & 0.2 $\pm$ 0.2 & 116 $\pm$ 22 & 194 $\pm$ 87 \\
\midrule
C/Qwen & 17.0 $\pm$ 6.8 & 0.0 $\pm$ 0.0 & 6 $\pm$ 2 & 166 $\pm$ 71 \\
C/Llama & 27.1 $\pm$ 2.8 & 0.2 $\pm$ 0.3 & 3 $\pm$ 1 & 218 $\pm$ 19 \\
C/OLMo & 10.2 $\pm$ 1.1 & 0.1 $\pm$ 0.1 & 2 $\pm$ 2 & 102 $\pm$ 15 \\
\midrule
Mean & 17.2 & 4.6 & & \\
95\% CI, cells fixed & [15.6, 18.8] & [3.3, 5.9] & & \\
Median & 18.7 & 0.2 & & \\
Mean w/o B/Qwen, C/Qwen & 16.8 & 5.9 & & \\
\bottomrule
\end{tabular}
\end{adjustbox}
\caption{Edits per cell as the cell value $\pm$ SD over three seeds: region removal (top; intervals with cells fixed condition on the nine cells and resample seeds for Kept and Acc.\ and problems for the differences) and the region graft at $\beta=2$ (bottom), whose recovered early-marker gap (\%) is a metric chosen after the grafts looped. $-$Top and $-$Cut give the accuracy gain (pp) over top removal and the length-matched cut, and the last row drops the two cells where the region beats top removal most.}
\label{tab:stats}
\label{tab:graftstats}
\end{table}

\begin{table}[!tbp]
\centering
\small
\setlength{\tabcolsep}{4pt}
\begin{adjustbox}{max width=\columnwidth}\begin{tabular}{lrrrrrrr}
\toprule
 & & \multicolumn{2}{c}{Dims} & \multicolumn{3}{c}{Coverage} & Two-seed \\
\cmidrule(lr){3-4}\cmidrule(lr){5-7}
Cell & Switch & Pre & Post & Pre & Post & Chance & (\%) \\
\midrule
A/Qwen & 0.4 & 5.4 & 1.3 & 0.414 & 0.050 & 0.018 & 56 \\
A/Llama & 0.4 & 6.2 & 3.3 & 0.431 & 0.084 & 0.007 & 55 \\
A/OLMo & 0.4 & 6.3 & 2.4 & 0.459 & 0.087 & 0.005 & 57 \\
\midrule
B/Qwen & 0.6 & 6.9 & 4.2 & 0.517 & 0.164 & 0.016 & 54 \\
B/Llama & 0.6 & 7.2 & 5.0 & 0.528 & 0.163 & 0.007 & 56 \\
B/OLMo & 0.6 & 7.7 & 5.5 & 0.548 & 0.179 & 0.005 & 56 \\
\midrule
C/Qwen & 0.6 & 6.7 & 1.6 & 0.530 & 0.141 & 0.018 & 57 \\
C/Llama & 0.6 & 7.3 & 1.3 & 0.542 & 0.142 & 0.007 & 58 \\
C/OLMo & 0.6 & 7.5 & 3.0 & 0.571 & 0.191 & 0.005 & 60 \\
\bottomrule
\end{tabular}
\end{adjustbox}
\caption{Switch $\lambda$, shared dimensions out of 16 before and after the switch (weighted by $\lVert \Delta W \rVert_F$), and coverage of the pre-switch region by pre-switch adapters (leave-one-out), post-switch adapters and chance. The last column is the share of its dimensions spanned by the two-seed $\lambda=0$ region at cut 0.75 (70--79\% at 0.65, 26--32\% at 0.85).}
\label{tab:5}
\label{tab:cuts}
\end{table}

\begin{table}[!tbp]
\centering
\small
\setlength{\tabcolsep}{2pt}
\begin{adjustbox}{max width=\columnwidth}\begin{tabular}{lrrrrrrrrr}
\toprule
 & \multicolumn{3}{c}{Pair A} & \multicolumn{3}{c}{Pair B} & \multicolumn{3}{c}{Pair C} \\
\cmidrule(lr){2-4}\cmidrule(lr){5-7}\cmidrule(lr){8-10}
 & Qwen & Llama & OLMo & Qwen & Llama & OLMo & Qwen & Llama & OLMo \\
\midrule
\rowcolor{band}\multicolumn{10}{l}{\emph{Early markers per 1,000 words at $\beta=2$}} \\
$\lambda=1$ & 0.16 & 0.13 & 0.14 & 1.54 & 0.83 & 1.19 & 0.06 & 0.07 & 0.10 \\
Region & 19.6 & 8.4 & 5.7 & 20.4 & 18.7 & 28.0 & 17.0 & 27.1 & 10.2 \\
Other & 0.41 & 1.03 & 0.29 & 1.12 & 3.44 & 0.80 & 0.16 & 0.23 & 0.08 \\
Whole & 12.1 & 8.3 & 10.7 & 11.9 & 70.4 & 59.0 & 1.7 & 16.9 & 10.2 \\
\midrule
\rowcolor{band}\multicolumn{10}{l}{\emph{Whole-response markers per 1,000 words}} \\
$\lambda=1$ & 0.48 & 0.12 & 0.41 & 4.68 & 2.77 & 5.14 & 0.07 & 0.14 & 0.33 \\
Region, $\beta=1$ & 1.72 & 1.06 & 0.82 & 6.92 & 5.88 & 20.82 & 1.72 & 0.81 & 0.47 \\
Other, $\beta=1$ & 0.24 & 0.51 & 0.32 & 2.94 & 7.25 & 4.64 & 0.14 & 0.35 & 0.29 \\
Whole, $\beta=1$ & 0.26 & 1.36 & 0.45 & 4.04 & 4.39 & 18.80 & 1.14 & 0.48 & 0.20 \\
Region, $\beta=2$ & 13.0 & 5.0 & 1.2 & 10.0 & 20.0 & 25.0 & 41.0 & 18.2 & 7.9 \\
Other, $\beta=2$ & 0.2 & 1.2 & 0.2 & 3.9 & 57.8 & 3.1 & 0.1 & 0.4 & 0.1 \\
Whole, $\beta=2$ & 29.2 & 11.3 & 5.1 & 6.5 & 113.4 & 39.0 & 8.8 & 27.1 & 9.5 \\
\midrule
\rowcolor{band}\multicolumn{10}{l}{\emph{Completed responses (\%)}} \\
Region, $\beta=1$ & 94 & 84 & 67 & 30 & 30 & 13 & 48 & 89 & 58 \\
Region, $\beta=2$ & 23.0 & 17.4 & 0.2 & 0.0 & 0.1 & 0.2 & 0.0 & 0.2 & 0.1 \\
Other, $\beta=2$ & 90.5 & 63.2 & 75.2 & 58.6 & 55.1 & 73.7 & 68.4 & 78.0 & 61.6 \\
Whole, $\beta=2$ & 30.3 & 23.3 & 1.8 & 8.6 & 0.5 & 0.0 & 0.0 & 6.2 & 0.4 \\
\midrule
\rowcolor{band}\multicolumn{10}{l}{\emph{MATH-500 accuracy (\%) at $\beta=2$}} \\
Region & 7.7 & 8.5 & 5.0 & 3.5 & 3.3 & 3.1 & 4.7 & 4.0 & 4.1 \\
Other & \textbf{28.9} & \textbf{18.8} & \textbf{10.2} & \textbf{19.2} & \textbf{13.2} & \textbf{9.5} & \textbf{17.8} & \textbf{24.8} & \textbf{9.7} \\
Whole & \underline{13.8} & \underline{12.1} & \underline{5.8} & \underline{7.5} & \underline{4.5} & \underline{3.2} & \underline{6.1} & \underline{7.2} & \underline{4.9} \\
\midrule
\rowcolor{band}\multicolumn{10}{l}{\emph{Tokens (thousands)}} \\
Region, $\beta=1$ & 1.1 & 2.0 & 1.7 & 5.8 & 5.8 & 3.5 & 4.3 & 1.5 & 1.9 \\
\midrule
\rowcolor{band}\multicolumn{10}{l}{\emph{Whole-response gap recovered (\%)}} \\
Region, $\beta=1$ & 6 & 3 & 3 & 16 & 7 & 134 & 10 & 3 & 1 \\
\midrule
\rowcolor{band}\multicolumn{10}{l}{\emph{Responses ending in a loop (\%)}} \\
$\lambda=0$ & 61 & 80 & 47 & 30 & 25 & 14 & 50 & 65 & 43 \\
$\lambda=1$ & 3 & 3 & 3 & 18 & 8 & 10 & 6 & 2 & 5 \\
Region removed & 8 & 7 & 1 & 3 & 3 & 1 & 4 & 4 & 1 \\
Region, $\beta=1$ & 4 & 7 & 26 & 55 & 50 & 80 & 16 & 4 & 24 \\
Region, $\beta=2$ & 71 & 69 & 96 & 84 & 90 & 90 & 80 & 90 & 92 \\
Other, $\beta=2$ & 6 & 27 & 18 & 33 & 34 & 18 & 14 & 9 & 11 \\
Whole, $\beta=2$ & 62 & 65 & 95 & 68 & 86 & 91 & 41 & 87 & 91 \\
\bottomrule
\end{tabular}
\end{adjustbox}
\caption{Grafts of the region, the same energy outside it (other) or the whole $\lambda=0$ update (whole) into the $\lambda=1$ adapter (seed means). Region and whole grafts rarely finish. A response loops when its last 300 words repeat at least half of their 10-grams. Bold and underline mark the two best accuracies.}
\label{tab:8}
\label{tab:9}
\label{tab:loop}
\label{tab:beta}
\end{table}

\end{document}